\documentclass[sigconf]{acmart}

\AtBeginDocument{%
  \providecommand\BibTeX{{%
    \normalfont B\kern-0.5em{\scshape i\kern-0.25em b}\kern-0.8em\TeX}}} 
    
\copyrightyear{2022}
\acmYear{2022}
\setcopyright{acmcopyright}\acmConference[MM '22]{Proceedings of the 30th ACM International Conference on Multimedia}{October 10--14, 2022}{Lisboa, Portugal}
\acmBooktitle{Proceedings of the 30th ACM International Conference on Multimedia (MM '22), October 10--14, 2022, Lisboa, Portugal}
\acmPrice{15.00}
\acmDOI{10.1145/3503161.3547972}
\acmISBN{978-1-4503-9203-7/22/10}

\usepackage{multirow}
\usepackage{adjustbox}
\usepackage{hyperref}

\begin{document}

\title{Everything is There in Latent Space: Attribute Editing and Attribute Style Manipulation by StyleGAN Latent Space Exploration}






\author{
	Rishubh Parihar\textsuperscript{1},
	Ankit Dhiman\textsuperscript{1,2}, 
	Tejan Karmali\textsuperscript{1},
	R. Venkatesh Babu\textsuperscript{1}
}
 \affiliation{
 	\textsuperscript{1}Indian Institute of Science, Bengaluru, \textsuperscript{2}Samsung Research 
 	\country{India}} 
\email{{rishubhp, ankitd, tejankarmali, venky}@iisc.ac.in} 








\begin{abstract}
Unconstrained Image generation with high realism is now possible using recent Generative Adversarial Networks (GANs). However, it is quite challenging to generate images with a given set of attributes. Recent methods use style-based GAN models to perform image editing by leveraging the semantic hierarchy present in the layers of the generator. We present Few-shot Latent-based Attribute Manipulation and Editing (FLAME), a simple yet effective framework to perform highly controlled image editing by latent space manipulation. Specifically, we estimate linear directions in the latent space (of a pre-trained StyleGAN) that controls semantic attributes in the generated image. In contrast to previous methods that either rely on large-scale attribute labeled datasets or attribute classifiers, FLAME uses minimal supervision of a few curated image pairs to estimate disentangled edit directions. FLAME can perform both individual and sequential edits with high precision on a diverse set of images while preserving identity. Further, we propose a novel task of Attribute Style Manipulation to generate diverse styles for attributes such as eyeglass and hair. We first encode a set of synthetic images of the same identity but having different attribute styles in the latent space to estimate an attribute style manifold. Sampling a new latent from this manifold will result in a new attribute style in the generated image. We propose a novel sampling method to sample latent from the manifold, enabling us to generate a diverse set of attribute styles beyond the styles present in the training set. FLAME can generate diverse attribute styles in a disentangled manner. We illustrate the superior performance of FLAME against previous image editing methods by extensive qualitative and quantitative comparisons. FLAME generalizes well on out-of-distribution images from art domain as well as on other datasets such as cars and churches.  \href{https://sites.google.com/view/flamelatentediting}{\color{blue} Project page}. 
\end{abstract}


\begin{teaserfigure}
  \includegraphics[width=\textwidth]{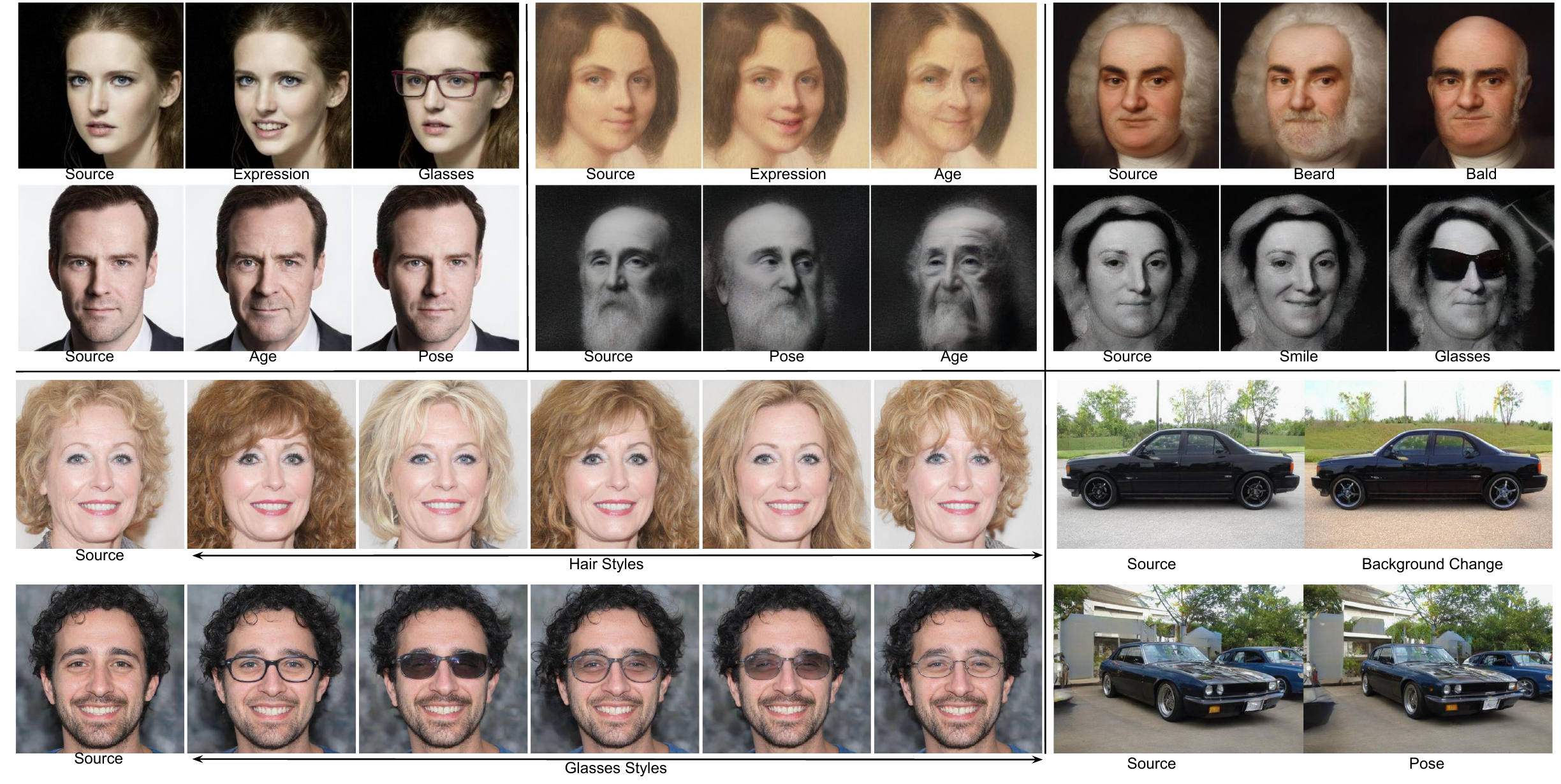}  
  \caption{ Examples of various attribute edits on synthetic faces and art images (Top). Example variations of attribute styles for eyeglasses and hairs generated by FLAME and edits on car dataset (Bottom).} 
  \label{fig:teaser} 
\end{teaserfigure}  

\maketitle 

\section{Introduction}
Image synthesis has been one of the long-standing problems in computer vision and graphics. With the advent of deep learning, many methods have been proposed for image synthesis in the last decade. Among all approaches, Generative Adversarial Networks~\cite{GAN-goodfellow} have shown promising results in generating photorealistic images. Recent GAN models such as StyleGAN~\cite{karras2019style,karras2020analyzing} architectures can generate images of diverse categories such as faces, cars, dogs, cats, churches, etc., that are often indistinguishable from natural images. Although these networks generate highly realistic images, it is challenging to control this generation process. StyleGAN architecture has layer-wise latent codes and stochastic vectors that control image generation. However, it requires additional methods to find disentangled latent transformations to generate images with given specifications. Semantic attribute editing in the latent space is a promising approach towards controlled image generation.  

The latent space of StyleGAN has rich semantic properties. Methods such as InterFaceGAN ~\cite{shen2020interpreting} and GANSpace~\cite{harkonen2020ganspace} demonstrate the existence of directions in latent space that controls the extent of attributes in the generated image. For example, there are directions that can manipulate camera pose, lighting, zoom level, and fine-grained facial attributes in the generated image. Prior works ~\cite{shen2020interpreting,harkonen2020ganspace,abdal2021styleflow,yuval2021agetransformation,latentclr,tewari2020pie,unsupervised_discovery,stylespace} estimate linear or non-linear paths in the latent space, achieving realistic attribute editing in StyleGAN generated images. GAN encoder models~\cite{psp,e4e,alaluf2021restyle,abdal2019image2stylegan} learn the mapping from the image space to the latent space to foster edits on real images. Once the image is mapped to its corresponding latent code, it can be edited in the latent space, just like synthetic images. However, the existing methods to estimate the attribute editing directions have two concerns, a) they require supervision from attribute classifiers trained on large data, and b) the estimated directions can be entangled with other attributes.

We propose a simple yet effective method to obtain disentangled attribute edit directions while using very less data: \textbf{F}ew-shot \textbf{L}atent-based \textbf{A}ttribute \textbf{M}anipulation and \textbf{E}diting (\textbf{FLAME}). 
FLAME is able to perform realistic edits for a wide variety of attributes - expression, pose, lighting, age, bangs, presence of glasses, hats, hair length, background change, day-night etc. (Fig. \ref{fig:teaser}). Our method requires minimal supervision of ten curated image pairs compared to previous methods requiring large-scale attribute annotations~\cite{shen2020interpreting} or pre-trained attribute classifiers~\cite{abdal2021styleflow,gao2021high,liang2021ssflow,yuval2021agetransformation}. Specifically, we create image pairs with a given attribute's presence and absence. We then compute the difference between the projected latent codes for the images in a pair. Finally, we estimate the dominant direction that aligns closely with these difference directions for all the image pairs in the dataset. Due to the attribute-specific image pairs, we obtain disentangled directions which change only one specific attribute while keeping other attributes unaffected. The estimated edit directions generalize well to diverse images and domains compared to previous works~\cite{abdal2021styleflow,gao2021high,liang2021ssflow,yuval2021agetransformation} that estimate instance-specific edits based on attribute scores. Further, we show (in Fig.~\ref{fig:teaser}) that directions obtained by FLAME from real images can be applied on out-of-domain artistic portraits. FLAME also generalizes to other popular categories such as cars and churches to obtain the attribute directions specific to the category under consideration.

While existing works find a direction for an attribute, they are not able to synthesize diversity within an attribute (for eg. diversity in hairstyles synthesized on a person) and is limited by the extent by which the direction is traversed. We propose a novel task of Attribute Style Manipulation (Fig. ~\ref{fig:teaser} Bottom), which aims to create diverse styles of a single attribute without changing other attributes and identity of the image. 
We propose a method that is a natural extension of our attribute editing framework to estimate the manifold of attribute styles in the latent space of a pre-trained StyleGAN. Sampling from this manifold generates images with variations in attribute styles keeping the identity and other image properties unchanged. We investigate our approach for attribute style manipulation for face images with two important face attributes: eyeglasses and hairstyle. This framework can have wide use in creating synthetic training datasets for training deep learning models for downstream applications. 

We summarize the main contributions of our work as follows:
\begin{itemize}
    \item We present a simple yet effective method \textbf{FLAME} that estimates disentangled linear directions in the latent space of StyleGAN using supervision from few ($\approx 10$) image pairs to perform highly realistic image edits.
    \item The directions estimated by FLAME generalize to out of domain art images and other categories: cars and churches. 
    \item To the best of our knowledge, we are the first to present a novel task of attribute style manipulation to generate diverse attribute styles, and demonstrate how FLAME can solve it. 
\end{itemize}

\section{Related Works}
\textbf{\textit{Image manipulation using GANs:}}
Recent style-based GAN architectures~\cite{karras2019style, karras2020analyzing} provide hierarchical control in the generated images ~\cite{yang2021semantic}. Multiple works ~\cite{shen2020interpreting, br2021photoapp,tewari2020stylerig,abdal2021styleflow,harkonen2020ganspace, gao2021high} perform fine-grained image editing by leveraging the rich structure present in the latent space of a pre-trained GAN. Another important direction of research involves training conditional GANs~\cite{mirza2014conditional} and cycle GANs~\cite{zhu2017unpaired} to perform image editing. MaskGAN~\cite{lee2020maskgan} learns a mapping between the segmentation mask and the rendered target to edit generated image. 
~\cite[]{wu2019relgan} conditions the image generator on attribute strengths to perform attribute edits. Although these methods can generate good quality image edits, they require retraining of the GAN model, which is computationally expensive for high resolution.
\begin{figure*}
  \centering
  \includegraphics[width=\linewidth]{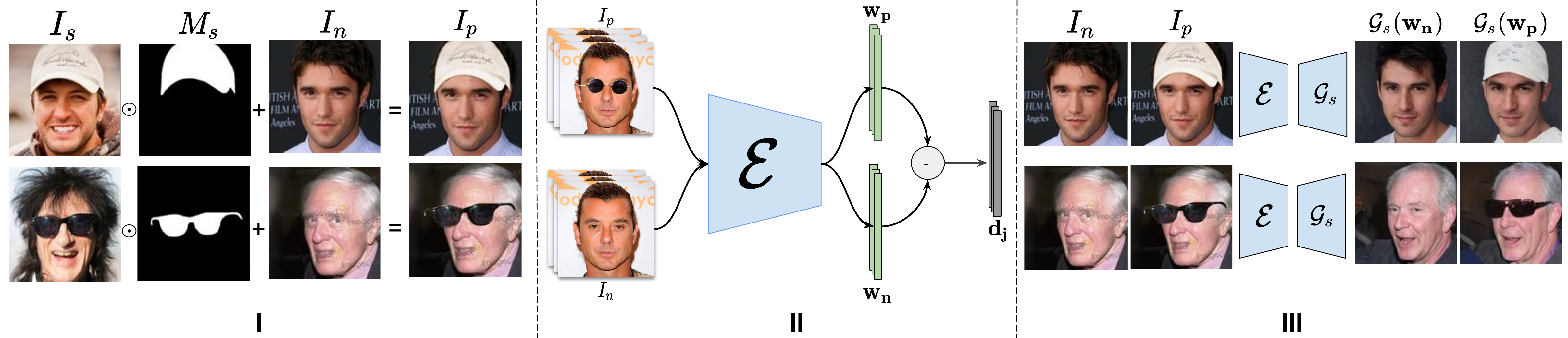} 
  \caption{~\textbf{Overview of our proposed method \textit{- I)}} Synthetic Pair Creation (Sec.~\ref{sec:synth_pair}). We create positive ($I_p$) and negative ($I_n$) images for an attribute $a_j$ using its mask. \textbf{\textit{II)}} Attribution direction is estimated from $\mathbf{d_j}$, which is difference of latent codes of $I_p$ and $I_n$ encoded by $\mathcal{E}$ into the latent space of StyleGAN2 (Sec.~\ref{sec:sem_dir_estimate}) \textbf{~\textit{II)}}. Reconstruction of $I_p$ and $I_n$ from the latent codes given by Encoder $\mathcal{E}$. Note that, although $I_p$ and $I_n$ do not look natural, the reconstructed pair looks more natural and has identity preserved across $\mathcal{G}_s(\mathbf{w_p})$ and $\mathcal{G}_s(\mathbf{w_n})$, where $\mathcal{G}_s$ denotes the synthesis network of the StyleGAN2 Generator $\mathcal{G}$.}
\label{fig:method_1}
\end{figure*}

\textbf{\textit{Image editing by latent-space manipulation}:}  
Recently, several strategies ~\cite{tewari2020pie, abdal2021styleflow, stylespace, tewari2020stylerig, abdal2020image2stylegan++} perform image editing by transforming $\mathcal{W/W+}$  latent space in a pre-trained StyleGAN model. Some works ~\cite{shen2020interpreting, harkonen2020ganspace, sefa, ling2021editgan} estimate global linear edit directions to model the latent transformation. In contrast, others learn a complex mapping that transforms the latent codes in an instance-specific manner for any given image ~\cite{abdal2021styleflow,yuval2021agetransformation,zhu2021barbershop,br2021photoapp}. One of the primary approaches is to estimate a linear direction of variation that controls any given attribute ~\cite{shen2020interpreting, sefa, harkonen2020ganspace, ling2021editgan} in a disentangled manner. It builds on the hypothesis that necessary semantic attributes are disentangled in the $\mathcal{W/W+}$ latent space~\cite{shen2020interpreting}. InterFaceGAN ~\cite{shen2020interpreting} trains a linear SVM in the latent space to estimate the attribute edit directions. In GANSpace ~\cite{harkonen2020ganspace}, a PCA is performed on latent codes to obtain the directions of maximum variations followed by manual filtering of directions. In SeFA ~\cite{sefa}, the author optimized for the latent directions such that the variations are maximized after projected on the affine matrix A. Further, multiple unsupervised methods ~\cite{latentclr, jahanian2019steerability, unsupervised_discovery, spingarn2020gan} discover latent transformations to perform editing. The above approaches can generate realistic attribute editing, but often entangle multiple attributes. 

EditGAN ~\cite{ling2021editgan} uses a pre-trained GAN ~\cite{datasetgan} that jointly models the image and segmentation mask to control the generated image using segmentation mask. It is shown in StyleSpace~\cite{stylespace} that the StyleSpace of the pre-trained StyleGAN model is more disentangled than other latent spaces. StyleRig ~\cite{tewari2020stylerig} leverage rich 3DMM models to create a mapping between StyleGAN and 3DMM semantics to obtain fine-grained control in the generated image. Ganalyze ~\cite{goetschalckx2019ganalyze} uses an assessor network to guide the discovery of the latent directions.


StyleFlow ~\cite{abdal2021styleflow} proposes continuous normalized flows to model the latent space transformations for a given attribute. Specifically, they use attribute classifiers to guide the training of the flow network. Similar to this, ~\citet[]{yuval2021agetransformation} learns a mapping in the latent space with the help of an age classifier. Thereon, many subsequent works leverage the attribute classifier for learning image edit operations ~\cite{gao2021high,khodadadeh2022latent,liang2021ssflow,yuval2021agetransformation,zhu2020improved-stylegan-what-are-good-latents}. However, the dependency of attribute classifiers limits the editing to a small set of attribute classes. Additionally, these classifiers increases the computation cost during when used during inference and training. Our proposed method can perform comparable edit operations without using any attribute classifier with minimal supervision. Like ours, PhotoApp ~\cite{br2021photoapp} trains a network to perform the latent transformation for lighting and head pose editing with limited supervision. StyleCLIP ~\cite{patashnik2021styleclip} performs text-driven image editing by leveraging the joint embedding between the image and the text. 

\textbf{\textit{GAN encoder models:}} 
GAN encoder models are used to learn mapping between real images to the latent space which can then be modified to perform image edits on real images. Multiple StyleGAN encoder models~\cite{psp,e4e,alaluf2021restyle,chai2021using,abdal2019image2stylegan,abdal2020image2stylegan++,xu2021continuity, alaluf2021hyperstyle} are proposed in the literature based on the use case of editability vs reconstruction. 
For StyleGAN models, the original $\mathcal{Z}$ space entangles multiple semantic concepts compared to the learned $\mathcal{W}$ space, which is more disentangled~\cite{karras2020analyzing}. Furthermore, $\mathcal{W+}$ space provides more flexibility as it allows separate latent codes for each generator layer. Most GAN encoder models map the input image to this immense $\mathcal{W+}$ space to obtain realistic reconstructions. Domain GAN inversion ~\cite{zhu2020domain} first performs inversion using an encoder followed by an optimization step which has a good reconstruction quality and is also semantically meaningful for editing tasks. PIE ~\cite{tewari2020pie} proposed a non-linear iterative optimization scheme to embed images in the latent space. ~\citet[]{xu2021continuity}proposes an encoder model for videos that uses optical flow.  ~\citet[]{chai2021using} trained the encoder with masked images, which results in the latent code corresponding to images while preserving the unmasked content in the input image.  


\section{Methodology}
In this section we present our method for estimating linear latent directions in the latent space of StyleGAN2 with few image pairs. StyleGAN2 generator $\mathcal{G}$ is composed of a mapping function $\mathcal{G}_m: \mathbb{R}^{512} \rightarrow \mathbb{R}^d$ and a synthesis function $\mathcal{G}_s: \mathbb{R}^{d} \rightarrow \mathbb{R}^{H \times W \times 3}$. Both the functions are represented as neural networks. Thus, $\mathcal{G} = \mathcal{G}_s \odot \mathcal{G}_m(\mathbf{z})$ where $\mathbf{z} \sim \mathcal{N}(\mathbf{0}^{512}, \mathbf{I}^{512 \times 512})$ (which is a normal distrbution with zero mean and identity covariance). $\mathcal{G}_m(\mathbf{z}) \in \mathcal{W+}$, which intermediate latent space of StyleGAN that offers disentanglement between different semantic concepts. Given this, we define a linear model for attribute editing as $\mathbf{w^\prime} = \mathbf{w_0} + \alpha\mathbf{d_j}$, where $\mathbf{w^\prime}, \mathbf{w_0} \in \mathcal{W+}$, $\mathbf{d_j} (\in \mathbb{R}^d)$ is the direction along which attribute $a_j$ changes; and $\alpha$ controls the strength of the change. For editing any attribute $a_j$, we curate a dataset $\mathcal{D}$ consisting of $n$ image pairs. We describe the dataset creation procedure in detail in Sec~\ref{sec:synth_pair}. Our hypothesis is that with image pairs that differ in only a single attribute $a_j$, we can estimate the direction along which $a_j$ changes. We demonstrate and validate this hypothesis in Sec.~\ref{sec:sem_dir_estimate}. Finally, we estimate directions for multiple styles of a single attribute and propose an algorithm to approximate the style manifold for that attribute in Sec.~\ref{sec:att-style}.

\subsection{Synthetic Pair Creation}
\label{sec:synth_pair}
For a given attribute $a_j$ we find a direction $\mathit{d_j} \in \mathcal{W+}$ such that traversing along $\mathit{d_j}$ alters only $a_j$ while keeping other attributes intact. To estimate the direction, we use a dataset consisting of $n$ image pairs $\mathcal{D}$. An image pair $\mathcal{D}^k = \{I^k_p, I^k_n\}$ has a positive image $I^k_p$, which contains the attribute $a_j$ and a negative image $I^k_n$ which does not contain $a_j$. Due to lack of availability of such paired datasets which have variation along a single attribute, we create a synthetic dataset which satisfies this property. 

We start by randomly sampling a set of $m$ negative images (${I^k_n}$) and $m$ source images (${I^k_s}$) (having $a_j$ present) with their part-wise segmentation mask (${M^k_s}$) from CelebAMask-HQ dataset~\cite{CelebAMask-HQ}. Thereafter, we use a simple cut and paste approach to create the corresponding positive image (${I^k_p}$). Specifically, given a $I^k_s$ and $M^k_s$, we choose the part-mask $M^k_s(j)$ that contains the regions corresponding to the attribute $a_j$. For example, mouth and hair region contains the attributes of expressions and bangs respectively. We blend $I^k_s$ and $I^k_n$ using $M^k_s(j)$ to obtain the positive image $I^k_p$ using Eq.~\ref{eq:2} and as shown in ~\ref{fig:method_1}-I). Note that we do not have to perform alignment of images ($I^k_n$ and $I^k_p$) as CelebAMask-HQ dataset has all eye-aligned images. The resulting positive image $I^k_p$ differs from the negative image $I^k_n$ only in x$a_j$; all other attributes are unchanged. Finally, from the generated pairs (\~ 30) image pairs $\mathcal{D}$, we manually select $10$ image pairs by discarding unnatural looking positive images. 
\vspace{-2mm}
\begin{equation}
\label{eq:2}
    I_p = (\textbf{1}-M_s) \odot I_n + M_s \odot I_s
\end{equation}  
\vspace{-2mm}


We create synthetic image pairs using this method for all the attributes located at different parts of the face (e.g., eyeglasses, smile, wearing-hat, adding-hair, bangs, facial hair, eye-close). However, not all the attributes can be transferred in this way and therefore we obtain positive-negative image pairs for these attributes in a different manner. We flip the negative image for the pose attribute to obtain the edited positive image. While for the age attribute, we use the framework proposed by SAM~\cite{yuval2021agetransformation} which is a state-of-the-art age editing method to obtain the "aged" version for a set of negative images. Additional details about pair creation are in the supplementary material. 

In Fig. ~\ref{fig:method_1}-I, it can be observed that the edited positive images for eyeglasses and hat attributes do not look realistic. However, the GAN encoder ~\cite{chai2021using} was trained on precisely such blended images, which maps them to a latent code decoded by the Generator. This Generative prior improves the realism of the input (positive/negative) images as shown in Fig. ~\ref{fig:method_1}-III. Although the mapped image pairs looked realistic, the inversion tends to change some subtle attributes during encoding. We propose an approach in Sec~\ref{sec:sem_dir_estimate} to estimate the directions which are robust to these variations. This proposed framework can obtain latent direction for any new edit operations by curating only a small pair of images, not possible with any previous method.

\subsection{Semantic Direction Estimation}
\label{sec:sem_dir_estimate}
We give an overview of our method for direction estimation in Fig.~\ref{fig:method_1}-II. As shown, our method relies on creation of attribute specific positive-negative image pairs, which we have described in Sec.~\ref{sec:synth_pair}. Having such a pair $D_k = (I^k_p, I^k_n)$ where $I^k_p$ and $I^k_n$ are the positive and the negative images respectively and $k \in \{1,2, \dots, n\}$, we project it into the $\mathcal{W+}$ latent space using StyleGAN2 encoder $\mathcal{E}$~\cite{chai2021using}. After projection, we obtain a dataset $L$ consisting of $n$ pairs of latent codes of the form $L_k=(\mathbf{w^k_p}, \mathbf{w^k_n})$, where $\mathbf{w^k_p} = \mathcal{E}(I^k_p)$ and $\mathbf{w^k_n} = \mathcal{E}(I^k_n)$. We then compute the difference direction for each latent pair as $\mathbf{d^k_j} = \mathbf{w^k_p} - \mathbf{w^k_n}$ and normalize it to unit length. Note that, all the $\mathbf{d^k_j}$ vectors correspond to the same attribute edit but from different image pairs. We want to estimate a direction $\mathbf{\hat{d_j}}$ which aligns closely with all of these difference vectors $\mathbf{d^k_j}$. To this end, we formulate an optimization problem to maximize the cosine similarity between $\mathbf{\hat{d_j}}$ and the difference vectors $\mathbf{d^k_j}$ as given in Eq.~\ref{eq:1}.

\begin{equation}
    \mathbf{\hat{d_j}} = argmax_{\mathbf{d_j}} \sum_{k=1}^{n}\left<\mathbf{d^k_j}, \mathbf{d_j}\right>^2
    \label{eq:1}
\end{equation} 

To solve the above optimization problem, we create a matrix $A$ by stacking all the $\mathbf{d^k_j}$ vectors as the rows. We then compute the Singular Value Decomposition of the matrix $\mathbf{A}$ to obtain: $\mathbf{A} = \mathbf{U}\mathbf{\Sigma}\mathbf{V^T}$. The column vector $\mathbf{v_*}$ of $\mathbf{V}$ matrix associated with the highest singular value will maximize the given optimization function (see supplementary material). One can also use the mean vector as the dominant direction, however we found SVD to perform better as it is more robust to outlier directions than mean vector. 

\subsection{Attribute Style Manipulation}
\label{sec:att-style} 

\begin{figure}[!t]
  \centering
  \includegraphics[width=0.70\linewidth]{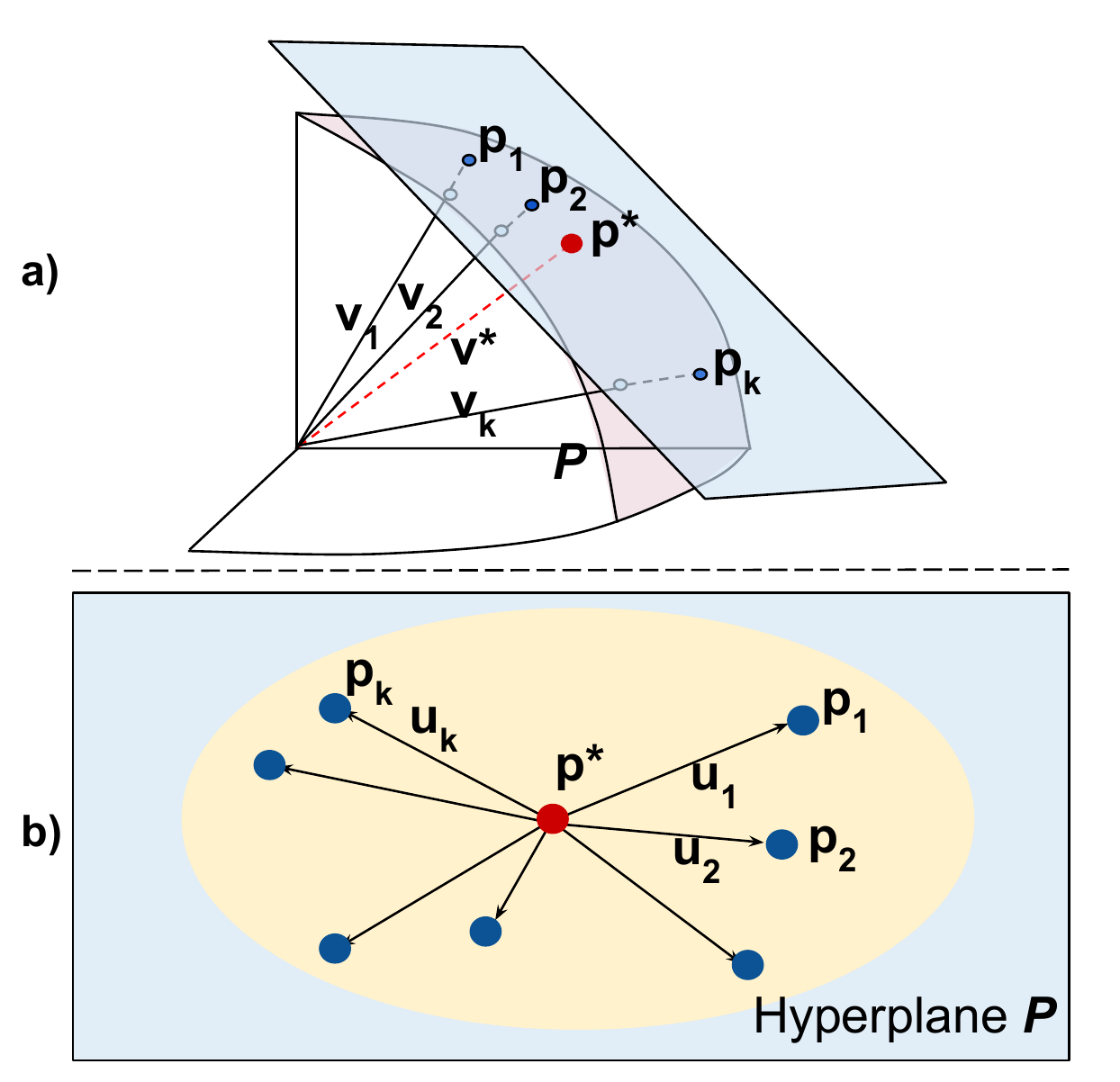} 
  \caption{~\textbf{\textit{a)}} Attribute Style Manifold: All the attribute style directions $\mathbf{v_k}$'s lie on the unit sphere (pink shaded) are projected onto the tangent hyperplane $P$ at $\mathbf{v*}$. \textbf{b)} Hyperplane $P$ where the primitive directions $\mathbf{u_k}$'s are estimated by taking a difference between $\mathbf{p_k}$'s and $\mathbf{p*}$}
\label{fig:method_2}
\end{figure}

We introduce a novel task of Attribute Style Manipulation and propose an algorithm to perform such attribute style edits with high fidelity. Current editing methods are limited to adding/removing any attribute or changing the attribute's strength such as age. However, for certain attributes such as hair, multiple styles exist, but the current method only alters the length of the hairs. To this end, we estimate a manifold for various styles for a given attribute to generate different styles. Images  having diverse styles of an attribute can be sampled from this manifold, while keeping the others unchanged. 

We sample $S$ positive images with different styles for the attribute $a_j$. We then estimate the direction for each style following the procedure given in ~\ref{sec:synth_pair} and ~\ref{sec:sem_dir_estimate}. We denoted the estimated directions for $S$ attribute styles (for $a_j$) as $\mathbf{v_k}$ for $k \in \{1,2,\hdots, S\}$ and use them to find a manifold for different styles. After this, we estimate the dominant direction $\mathbf{v^*}$ that aligns with all the normalized $\mathbf{v_k}$'s by solving the optimization problem similar to Eq.~\ref{eq:1}. As the $\mathbf{v_k}$'s and $\mathbf{v^*}$ are normalized to unit length, we shift them to origin so that they lie on the surface of a unit sphere as shown in Fig. ~\ref{fig:method_2}-a. To find a new attribute style, we wish to sample vectors on the surface of this sphere in the neighborhood of $\mathbf{v_k}$'s. However, it is challenging to directly sample a vector from the desired region on the sphere.

To this end, we first compute a tangent hyperplane $P$ to the sphere at point $\mathbf{v^*}$ and extent all the $\mathbf{v_k}$ vectors up to $P$ to obtain the intersection points $\mathbf{p_k}$'s and ${p^*}$(=$\mathbf{v^*}$) as shown in Fig.~\ref{fig:method_2}-a. To sample a point on the sphere, we can sample a point on the hyperplane $P$ and then project it back onto the sphere's surface by normalizing it to a unit length. Hence, we estimate the primitive vectors $\mathbf{u_k}$'s lying on $P$ using Eq.~\ref{eq:3} by subtracting the intersections $\mathbf{p_k}$’s from ${\mathbf{p^*}}$ as shown in Fig.~\ref{fig:method_2}-b. We take a linear combination of $\mathbf{u_k}$'s to sample a point $\mathbf{b}$ on the hyperplane $P$ as given in Eq.~\ref{eq:4}. $\mathbf{b}$ is then projected back onto the unit sphere by normalizing it and thus, it is now a new sampled point on the sphere surface. Note that we wish to sample from the neighborhood of the vectors $\mathbf{v_j}$ hence we sample small values for the weights $\lambda_i$ from the range of $(-\epsilon, \epsilon)$. Finally, for any desired image $I$ for which attribute style variation is to generated, we first project it to $\mathcal{W+}$ as $\mathbf{w} = \mathcal{E}(I)$, and manipulate it as $\mathbf{w^\prime} = \mathbf{w} + \alpha \mathbf{b}$. We explore other method to sample $\mathbf{b}$ - convex combination of $mathbf{u_k}$ and modifyeing the attribute strengths $mathbf{\alpha}$ in the supplementary material.

\begin{equation}
\label{eq:3}
    \mathbf{u_k} = \mathbf{p_k} - \mathbf{p^*}   \;\;\;\;\;\; k \in \{1, 2, \ldots S\}
\end{equation}
\begin{equation}
\centering
\label{eq:4}
    \mathbf{b} = \sum_{k=1}^S \lambda_k \mathbf{u_k}
\end{equation}

\section{Experiments}
This section will present the results and experiments to evaluate our method for attribute editing and attribute style manipulation. We use CelebAMask-HQ ~\cite{CelebAMask-HQ} dataset and test set of StyleFlow ~\cite{abdal2021styleflow} for all of our experiments on face images. For art images we used Metfaces dataset ~\cite{Metfaces}, LSUN cars ~\cite{lsun} for cars and LSUN church ~\cite{lsun} for churches.
For creating the synthetic dataset as explained in Sec. ~\ref{sec:synth_pair}, we used segmentation mask from CelebAMask-HQ ~\cite{CelebAMask-HQ} along with the attribute labels from ~\cite{liu2015faceattributes}. We use StyleGAN2~\cite{karras2020analyzing} model, trained on facial images to generate images and a pre-trained encoder from ~\cite{chai2021using} for mapping real images to latent codes. 

Building from the intuition that each layer of StyleGAN Generator controls different hierarchical properties~\cite{yang2021semantic}, we define a set of layers for each attribute editing as follows: for hair and hat 0-6, eyeglasses 0-9, smile 5-6, pose 0-4, facial hair 6,7 and 10, lighting 7-18 and eye-close 5-7. We have empirically found that modifying only the above-selected layer for editing any attribute performs the best. This is not uncommon practice to alter only few layers for editing of any given attribute and all the state-of-the-art methods follow this approach ~\cite{abdal2021styleflow, harkonen2020ganspace, latentclr}.   

\begin{figure*}[h!]
  \centering
  \includegraphics[width=\linewidth]{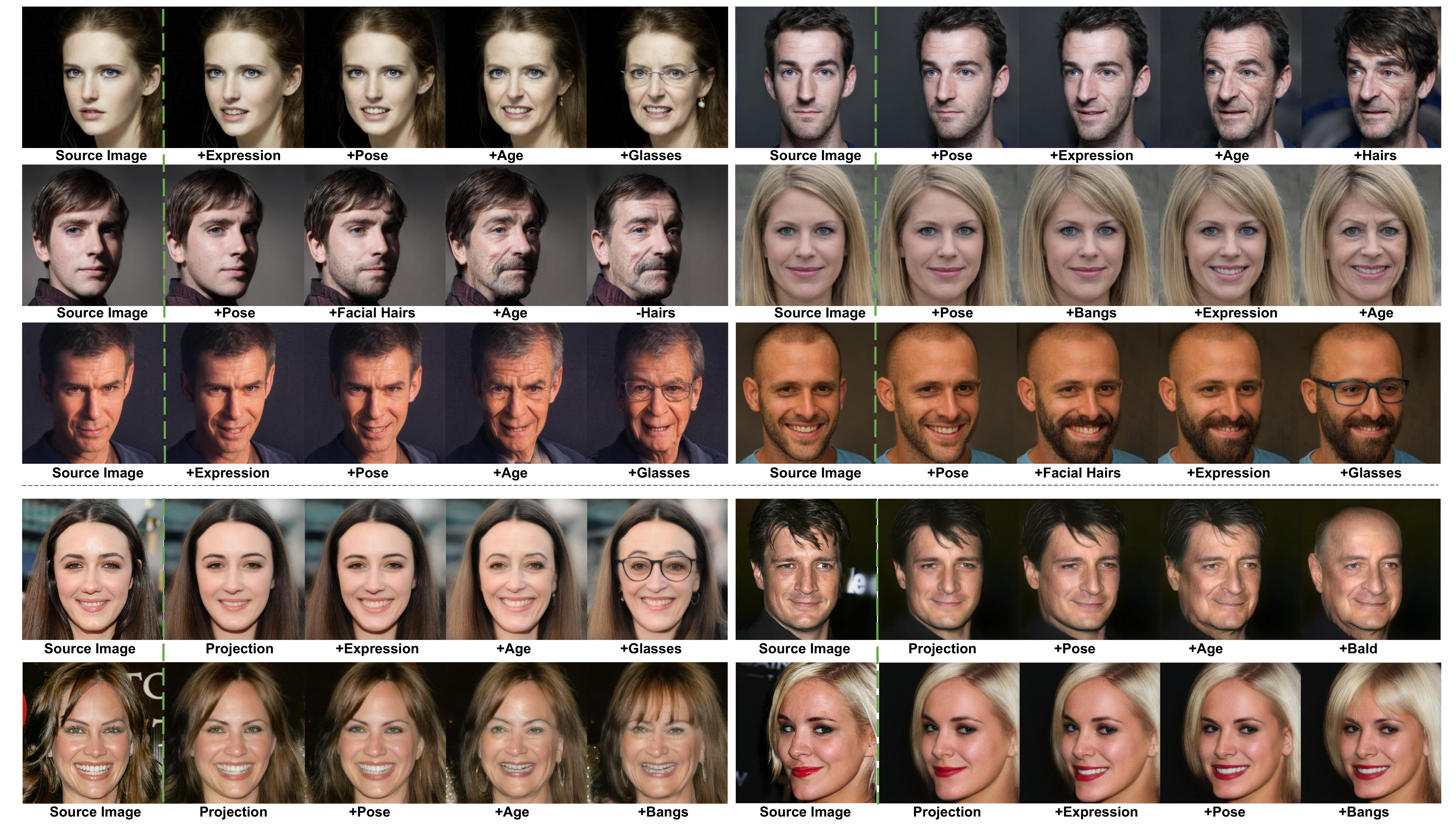}  
  \vspace{-5mm}
  \caption{~\textbf{Results for sequential attribute editing on synthetic-images (Top). Sequential attribute editing on real images (Bottom). }} 
  \label{fig:results1-synth} 
\end{figure*}

\begin{figure*}[h!]
  \centering
  \includegraphics[width=\linewidth]{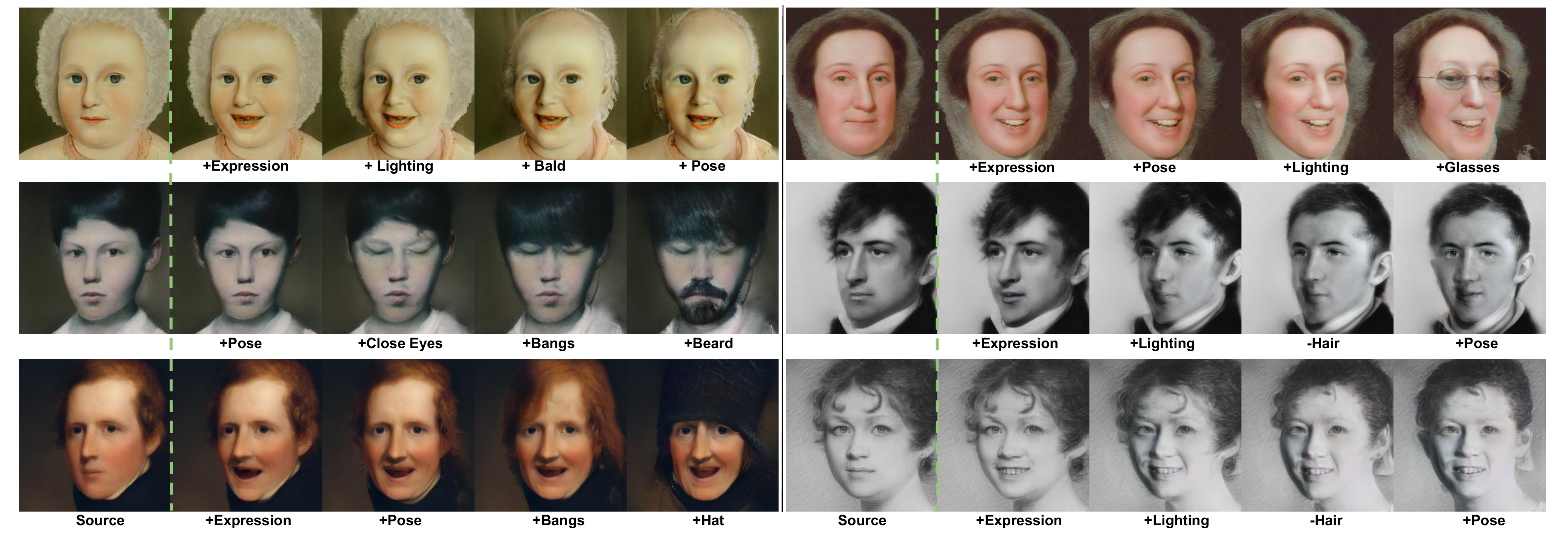} 
  \caption{~\textbf{Results for sequential attribute editing on out-of-domain images from MetFaces dataset. Out-of-domain results with such high fidelity is not possible by any of the previous methods}} 
  \label{fig:results1-metfaces} 
\end{figure*}

\begin{figure}[h!] 
  \centering
  \includegraphics[width=\linewidth]{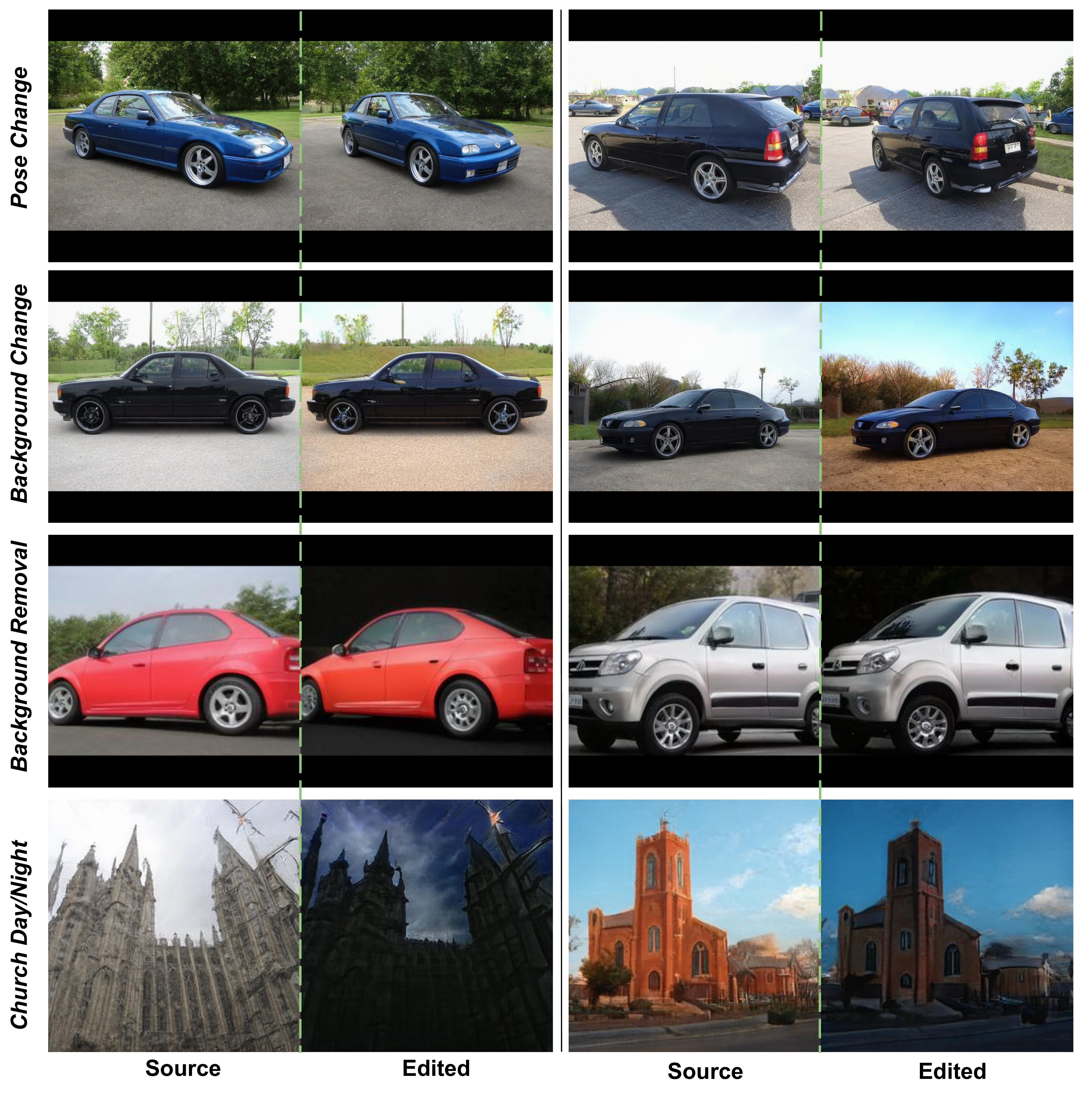} 
  \caption{~\textbf{Attribute editing on cars and church datasets}} 
  \label{fig:results_car_church} 
\vspace{-3mm}
\end{figure}

\subsection{Attribute Editing} 
We show results for a diverse set of face images and Out-of-Domain (OOD) art images from Metfaces edited using random sequential attribute editing in Fig. ~\ref{fig:results1-synth} and Fig.  ~\ref{fig:results1-metfaces} respectively. The edited images from our approach look realistic and coherent even though we use only ten synthetic image pairs. We observe that the edited images closely resemble the original image, maintaining a person's identity. Also, note that while editing any attribute, all the other attributes are unchanged, proving that our edit directions are largely disentangled. Additionally, FLAME does not modify the background and the skin tone during the edits. Interestingly for art images, FLAME is able to preserve the identity and the painting style during the editing. Note that, the edit directions are obtained from the real image pairs and they do generalize really well on art images. Additionally, we have also performed editing on real face images by first encoding the input image into the W+ latent space using encoder \cite{chai2021using} and using the obtained latent code for editing. Results for real image editing is shown in Fig. ~\ref{fig:results1-synth} (Bottom). One can observe that FLAME results in realistic attribute editing on real images and the edits are disentangled. 
    
\textbf{Ablation on image pair selection:} To evaluate the robustness of our method against the pairs selected for direction estimation, we perform an experiment with $5$ novel expert volunteers. The volunteers were asked to select the most natural looking $10$ image pairs ($D^k$) from generated $100$ image pairs. We then estimate the dominant direction ($\hat{d_j}$) for each of these set of image pairs as explained in Sec.~\ref{sec:sem_dir_estimate} and computed the pair-wise cosine similarity between them. Tab.~\ref{tab:random-volunteers} shows the histogram of similarity scores. We can observe that most of the directions are highly correlated as evident in the distribution which is skewed towards large values. This suggests that a new user can easily create the required image pairs with minimal efforts to find edit directions and our method is robust to the choice of specific image pairs used.  

\begin{table}[]
\caption{Distribution of pair-wise cosine similarity between directions obtained by multiple image-pair sets selected by novel volunteers. The statistics is aggregated over three attributes: pose, age and eyeglass.}
\begin{adjustbox}{max width=\linewidth} 
\begin{tabular}{l|llll|l}
\hline
Cosine Similarity $\uparrow$ & $0.0-0.7$ & $0.7-0.8$ & $0.8-0.9$ & $0.9-1.0$ & Mean $\uparrow$ \\ \hline
Normalized Frequency & $0$ & $0.133$   & \textbf{$0.300$} &   \textbf{$0.567$}   & \textbf{$0.893$} \\ \hline
\end{tabular}
\end{adjustbox}
\label{tab:random-volunteers}
\end{table}

\vspace{-4mm}
\subsection{Comparison with state-of-the-art methods}
We compare FLAME quantitatively and quantitatively  with three recent face editing methods - InterFaceGAN~\cite{shen2020interpreting}, GANSpace~\cite{harkonen2020ganspace} and StyleFlow~\cite{abdal2021styleflow}. InteFaceGAN and StyleFlow are supervised methods, whereas GANSpace is an unsupervised method. For InteFaceGAN, we use latent directions for expression, pose, age and eyeglass attributes from the provided implementation on StyleGAN2~\cite{karras2020analyzing}. We use the implementation provided by the authors for GANSpace to estimate the PCA and manually select those principal components which correlate with expression, pose, age and eyeglass attributes. For StyleFlow, we use the original codebase for editing images for the above set of attributes.
In StyleFlow and GANSpace original implementation only a subset of layers is modified for editing and InterFaceGAN modify all the layers as they train a SVM. We have kept the same configuration during this experiment for a fair comparison. In this experiment, we estimate the attribute edit directions with $10$ synthetic image pairs. We use the test set of StyleFlow for evaluation purposes as any of the methods did not use it during training. For comparison, we perform individual and sequential edits for expression, pose and age attribute editing as these are the common attributes in all four methods.

\begin{figure*}[h!]
  \centering
  \includegraphics[width=\linewidth]{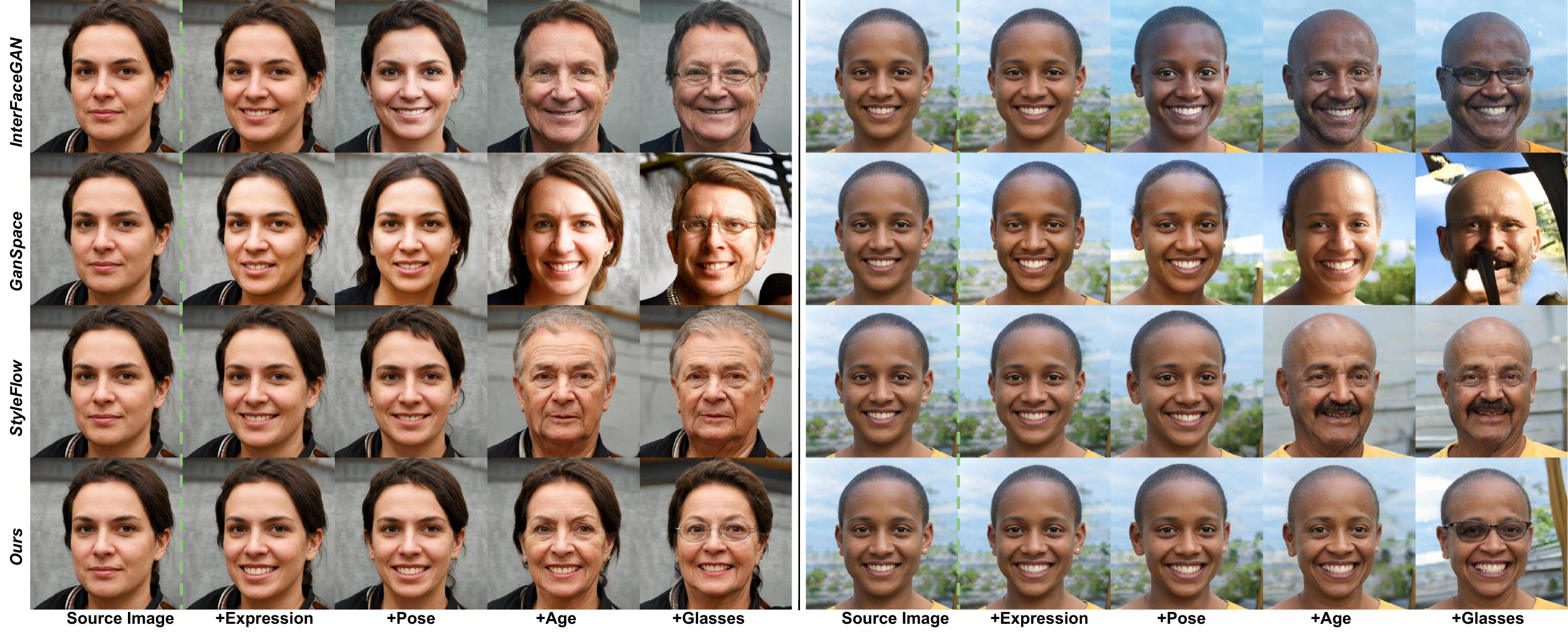}  
  \caption{~\textbf{Qualitative comparison for sequential attribute edits on synthetic face images with following attribute editing approaches: InterfaceGAN ~\cite{shen2020interpreting}, GANSpace~\cite{harkonen2020ganspace}, StyleFlow~\cite{abdal2021styleflow}}}
  \label{fig:comparison-results}
\end{figure*}

\textbf{\textit{Qualitative Comparison:}}
We compare FLAME against other methods by performing sequential image edits by iteratively changing attributes using the sequence: expression, pose, age, and eyeglasses. Fig. ~\ref{fig:comparison-results} shows the visual results for sequential edits. We observe that our method retains the identity and face structure well even after multiple edit operations. InterFaceGAN and StyleFlow erroneously change the gender while editing the age attribute in both the examples and GANSpace alters the gender while adding glasses. Note that GANSpace entangles multiple attributes, inducing a change in lighting and skin tone. StyleFlow generates realistic edits and doesn't change most attributes but alters identity after a few sequential edits.

\begin{figure*}[h!]
  \centering
  \includegraphics[width=\linewidth]{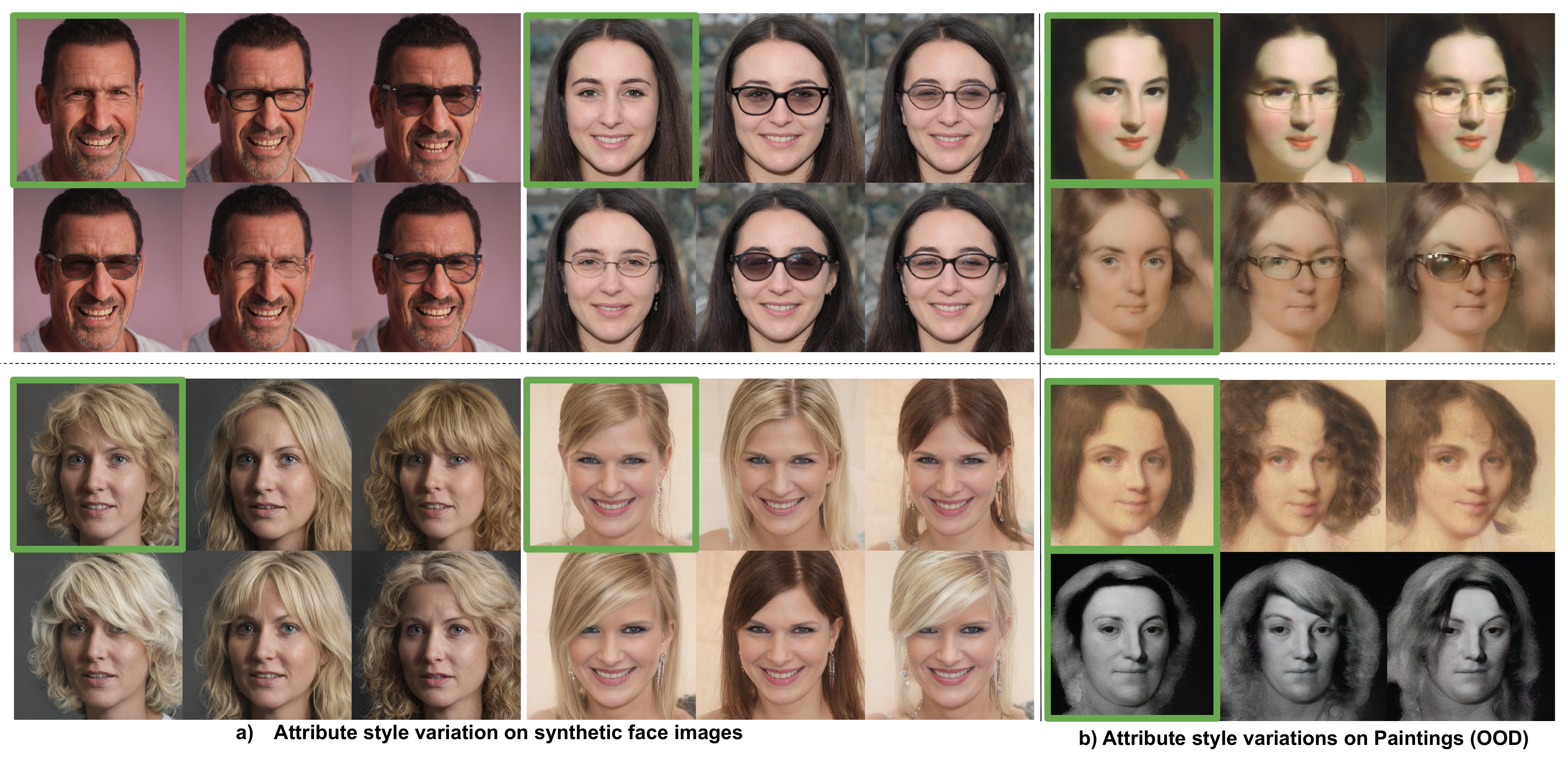}   
  \vspace{-5mm}
  \caption{~\textbf{Results for glasses and hair-style attribute variations generated for a given synthetic face images and Out-of-Distribution art images. Source image is shown in green inset, zoom-in for better viewing.}}
  \label{fig:attribute}
\end{figure*}  

\textbf{\textit{Quantitative Comparison:}}
We compare the FID (Fr\`echet Inception Distance) scores of the sequentially edited images and individual edited images from all four methods to quantify the quality of edits. For sequential editing, we applied the following edit sequence: \textit{expression, pose, age} and used the final edited image for comparison. For individual editing, we used separately edited image for each of the above three attributes. We follow the experimental setup from~\cite{abdal2021styleflow} and compute the FID score, we use 1k samples generated by a pre-trained StyleGAN2 model with a truncation factor of 0.7. Tab. ~\ref{tab:sequential-comparison}. and Tab. ~\ref{tab:individual-comparison} presents results of this experiment. Our method achieves the lowest FID score among all the methods, demonstrating that our generated edits have a high realism even after performing multiple sequential edits. We also compare our method for identity preservation as it is an essential metric for evaluating face editing algorithms. We use a state-of-the-art face recognition model ~\cite{face-rec-2020} to obtain the face embeddings for the original and sequentially edited image and compute the Cosine Similarity (CS) and Euclidean Distance (ED) to quantify the identity preservation. Our method performs at par with the best performing GANSpace~\cite{harkonen2020ganspace} method on both CS and ED metrics. However, we observe from Fig. ~\ref{fig:comparison-results} (second row) that the age editing direction for GANSpace is not disentangled and is incapable of performing significant changes in age. This results in an age-edited image similar to the original identity.

\setlength{\tabcolsep}{4pt}
\begin{table} 
\begin{center}  
\caption{\textbf{Comparison for sequential image editing (expression, pose, age).}}  
\label{tab:sequential-comparison} 
\begin{tabular}{ccccc}  
\hline\noalign{\smallskip}
Methods & FID $\downarrow$ & ED $\downarrow$ & CS $\uparrow$ & User Study $\uparrow$ \\
\noalign{\smallskip}
\hline
\noalign{\smallskip} 
InterFaceGAN &  43.07  &  0.61  & 0.92 & 20.40 \\
GANSpace  &  42.38  & \textbf{0.50}  & \textbf{0.95} & 8.70 \\
StyleFlow  &  47.81  &  0.71  & 0.82  & 16.31 \\
FLAME     &  ~\textbf{34.59}  & \textbf{0.50}  & \textbf{0.95} & ~\textbf{54.59} \\ 
\hline
\end{tabular}
\end{center} 
\end{table} 
\setlength{\tabcolsep}{1.4pt}

\setlength{\tabcolsep}{4pt}
\begin{table}
\begin{center} 
\caption{Comparison for individual attribute editing}
\label{tab:individual-comparison}
\begin{adjustbox}{max width=\linewidth}
\begin{tabular}{cccccc}
\hline 
Attribute  & Metric & InterfaceGAN & GANSpace & StyleFlow & FLAME  \\
\hline 
\multirow{3}{*}{Expression} & FID $\downarrow$   & 36.45        & 36.32    & 34.01     & \textbf{33.98}  \\
      & CS  $\uparrow$   & 0.98       & 0.98   & 0.99    & \textbf{1.00} \\
      & ED  $\downarrow$   & 0.32        & 0.29    & 0.23     & \textbf{0.15}  \\
      \hline
\multirow{3}{*}{Pose}  & FID $\downarrow$  & 34.53        & 34.51    & 34.34     & \textbf{30.81}  \\
      & CS  $\uparrow$   & 0.97       & 0.97   & 0.97    & \textbf{0.98} \\
      & ED  $\downarrow$   & 0.38        &  0.36    & 0.36     & \textbf{0.28}  \\
      \hline
\multirow{3}{*}{Age}   & FID $\downarrow$   & 36.69        & 36.24    & 47.82     & \textbf{34.11}  \\
      & CS $\uparrow$     & 0.93       & \textbf{0.95}   & 0.89    & \textbf{0.95} \\
      & ED $\downarrow$    & 0.55        & \textbf{0.47}    & 0.70     & 0.48 \\ 
\hline
\end{tabular}
\end{adjustbox}
\end{center}
\end{table}
\setlength{\tabcolsep}{1.4pt}  

\textit{\textbf{User Study:}}
We conducted a user study to compare FLAME with InterFaceGAN, GANSpace, StyleFlow, in which $24$ images were presented to $25$ participants. The participants were shown the original image and the final sequentially edited image along with intermediates edited images in the sequence for the four methods in random order. The volunteers were asked to select the best editing results based on identity preservation and overall visual quality. We use the following sequence of operations expression, pose, age and eyeglasses to generate editing results. Tab. ~\ref{tab:sequential-comparison}. compiles the results from this user study and shows that FLAME was selected most of the time (54.59\%) followed by InterFaceGAN (20.40\%), StyleFlow (16.31\%) and GANSpace (8.70\%). 

\textit{\textbf{Qualitative Results on Car and Church categories.}} To show the generalization ability of our proposed method, we performed image editing on two additional datasets of cars and churches.
For cars, we performed three edits: pose-change, background-change and background removal as shown in top three rows in Fig. \ref{fig:results_car_church}. For churches, we performed day-to-night editing which is shown in the bottom row in Fig. \ref{fig:results_car_church}. We use ten curated image pairs (See pairs in supplementary material) and pre-trained StyleGAN encoder models for cars and churches provided by \cite{chai2021using}. For cars, our method preserves all the fine details such as the orientation of wheel-rim, color, head and tail lights in the pose and background change tasks for cars. Similarly, it preserves the structure for day-night editing for churches. The wheel rim has changed for the background removal edit in cars, but all other fine details are unchanged. These results substantiate that our approach works effectively for other classes. 
\vspace{-5mm}
\subsection{Attribute Style Manipulation}
\label{results-att-style}
Fig.~\ref{fig:attribute} presents the generated diverse style variations for hair and eyeglass attributes. We empirically found the following values for hyper-parameters works best: $\lambda_i \in (-0.35, 0.35)$, and edit strength $\alpha \in (0.36, 0.46)$ and $\alpha \in (0.48,0.58)$ for eyeglass and hair, respectively. As shown in Fig.~\ref{fig:attribute}, our methods can generate diverse frame shapes ranging from frameless to big frames for eyeglasses. The generated results also include sunglasses with varying transparency in the lens. Similarly, our method generates diverse structures and appearances for hairstyles, as shown in Fig.~\ref{fig:attribute}. Observe that, in the third original image, the forehead was partially hidden by the hair. Still, new hairstyles are generated in some of the generated images where the forehead is completely visible.

All the generated attributes styles look realistic and match the face and the image's background well. Note that most of the other image properties like identity are unchanged during style manipulation, while lighting and background do not change significantly. However, there are very subtle changes in expressions but are majorly unnoticeable.

We compare the embeddings from the face-recognition network~\cite{face-rec-2020} to quantitatively evaluate the identity preservation in the generated samples. We generated $100$ attribute style variations for six sets of images for both eyeglass and hair. Then, we computed the CS and ED between the original and style-edited image. We obtained a CS score of $0.976$ and ED score of $0.34$, and for eyeglass, we obtained a CS score of $0.956$ and ED score of $0.457$. These results imply that our method well preserves identity in generated images. 

\vspace{-4mm}
\section{Discussion and Conclusion} 
In this work, we propose a simple yet effective approach FLAME for face attribute editing by discovering disentangled linear directions in the latent space of the pre-trained StyleGAN model. Our method requires only a few synthesized image pairs to obtain attribute edit directions. We show extensive results for both qualitatively and quantitatively for our method. One limitation of our work is curating synthetic image pairs which can be difficult in some cases, such as gender editing. Similar to existing editing works, our method can also be potentially misused for malicious purposed. Additionally, we propose a novel method to generate attribute style variations for glasses and hairstyles, keeping other attributes unchanged. The proposed framework of attribute style manipulation can be used to generate synthetic image datasets for multiple downstream tasks.

\vspace{1mm} \noindent \textbf{Acknowledgements.}
Rishubh Parihar acknowledges the support from Prime Minister’s Research Fellowship (PMRF). 


\bibliographystyle{ACM-Reference-Format}
\bibliography{egbib}

\clearpage

\appendix 

\begin{center}
\textbf{\LARGE Supplementary Material - Everything is There in Latent Space: Attribute Editing and Attribute Style Manipulation by StyleGAN Latent Space Exploration}
\end{center}



\section{Introduction} In the main paper, we demonstrated the effectiveness of FLAME for attribute manipulation. In this document, we provide additional details and ablations for FLAME. We first evaluate our attribute style manipulation algorithm~\ref{sec:eval}, followed by explanation on the pair creation~\ref{sec:synth_pair}. Finally, we show additional results on churches, cars and face dataset for both attribute editing and attribute style manipulation in later sections. Please find results for interpolation between edits and diverse attribute styles in the supplementary video. 

\section{Evaluation for attribute style manipulation} 
\label{sec:eval}
To evaluate our proposed algorithm for diverse attribute style generation, we compared FLAME against two baseline approaches for Attribute Style Manipulation.

\textbf{Baseline 1}:
Changing the strength of attribute editing can also
generate diverse attribute styles. For example, changing the strength of glasses attribute transforms the transparent eyeglasses into big dark-colored sunglasses. Specifically, to generate various attribute styles, we sample the edit strength $\alpha$ from a given range  $(l,r)$,, to scale the average dominant direction vector $\mathbf{v^*}$ and transform the latent code $\mathbf{w}$ as follows: 

\begin{equation}
    \mathbf{w^\prime} = \mathbf{w} + \alpha \mathbf{v^*} 
\end{equation}

\textbf{Baseline 2}: 
For the second baseline, we take a convex combination
of the primitive directions $u_k's$ to sample a new point from the style manifold as follows:

\begin{equation}
    \mathbf{b} = \sum_{k=1}^S \lambda_k \mathbf{u_k}
\end{equation}

\begin{equation}
    \sum_{k=1}^S \lambda_k = 1  \;\;\;    and \;  \lambda_i > 0
\end{equation}

\begin{figure}[h!]
\includegraphics[width=0.5\textwidth]{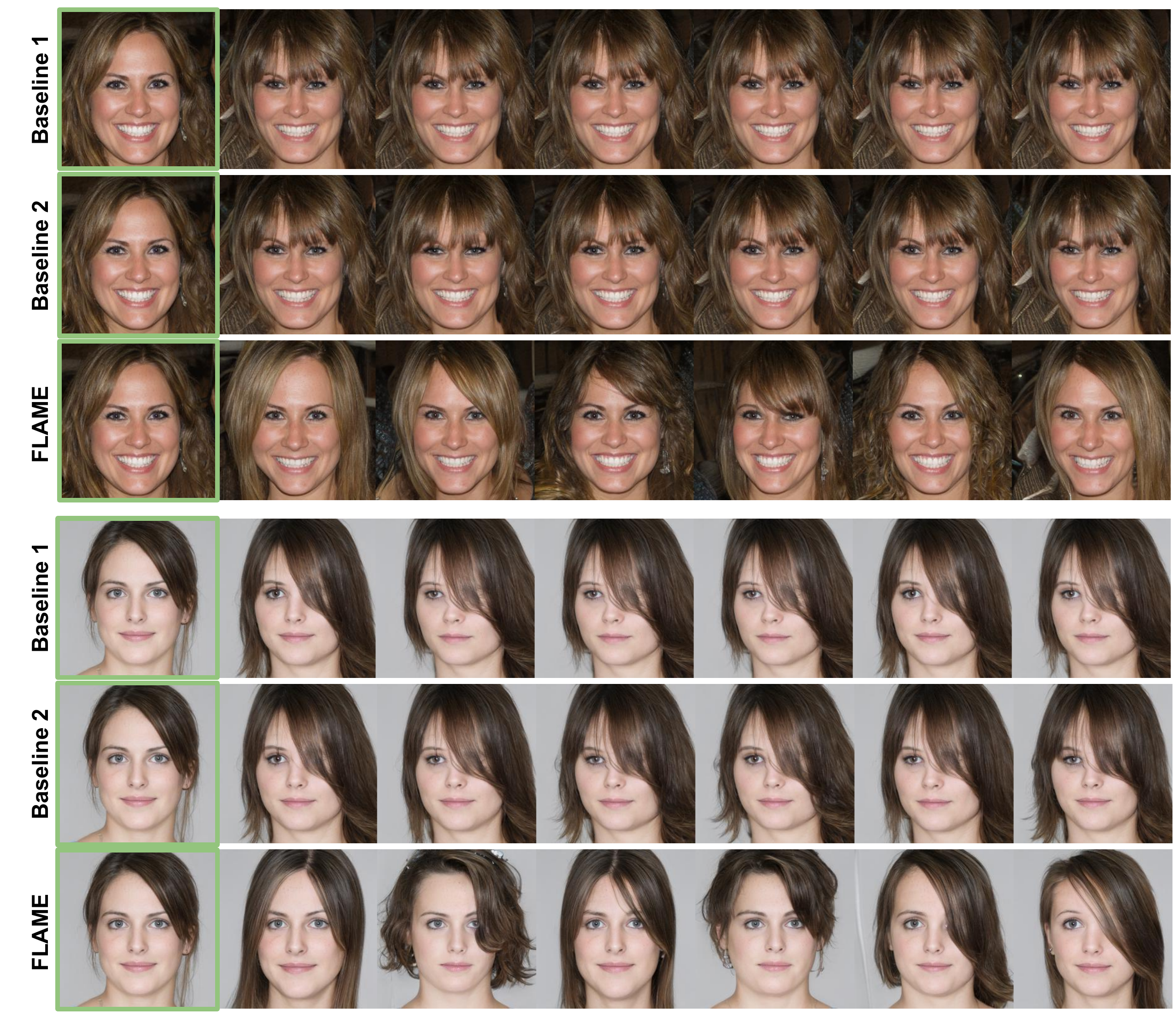} 
\caption{Evaluation of FLAME for hair style manipulation with baselines. Source image is given in green inset.}  
\label{fig:hair-eval-attr-style}
\end{figure}

\begin{figure}[h!]
\includegraphics[width=0.5\textwidth]{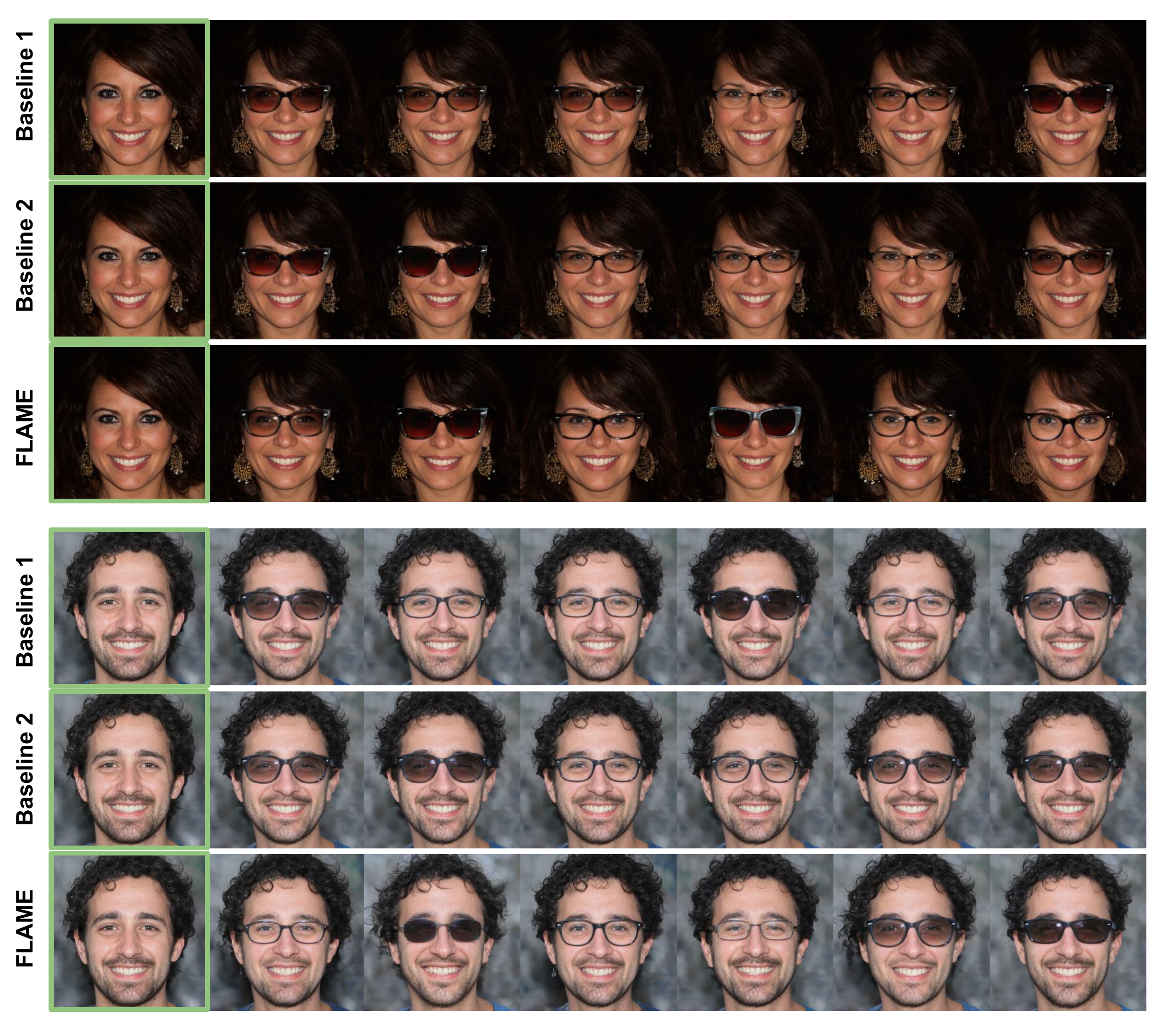} 
\caption{Evaluation of FLAME for eyeglass style manipulation with baselines. Source image is given in green inset.}  
\label{fig:eye-glass-attr-style} 
\end{figure}

Note that the above formulation is limited to interpolation between
the attribute styles corresponding to the primitive vectors $\mathbf{u_k}$. In contrast, our proposed algorithm can extrapolate beyond the primitive vectors by taking values of $\lambda_k < 0$. Some of the qualitative results of this experiment are shown in Fig.~\ref{fig:hair-eval-attr-style} and Fig.~\ref{fig:eye-glass-attr-style}. It can be observed that FLAME generates diverse styles of hair and eyeglass. On the other hand, both baselines collapse to only a few styles. This suggests that naively changing the attribute strength is insufficient to generate diverse attribute styles. In the eyeglass examples, Baseline 2 works better than Baseline 1, but both of them struggle to generate diverse outputs for hairstyle attributes.

\section{Method for Synthetic Pair Creation} 
\label{sec:synth_pair}
\begin{figure}[h!]
    \centering
    \includegraphics[width=0.9\linewidth]{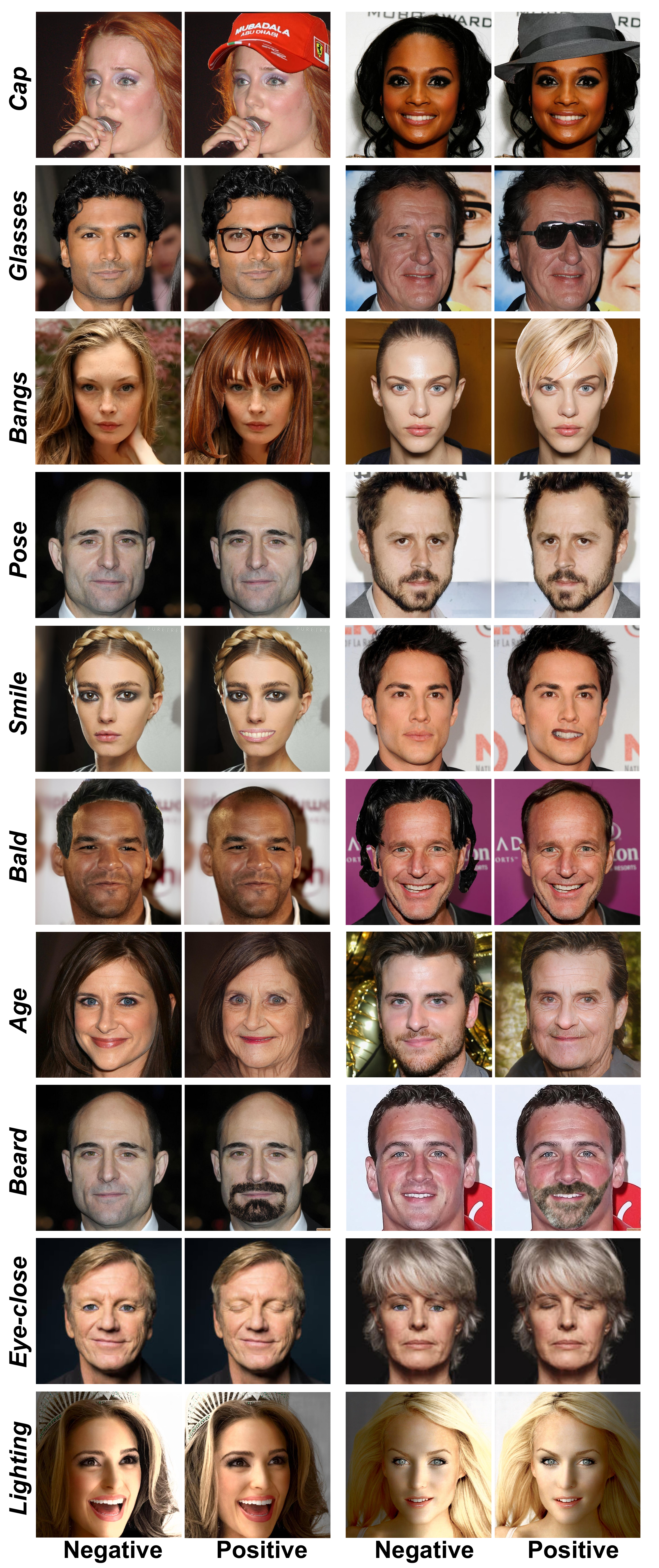}
    \caption{Sample image pairs for face attribute edits.}
    \label{fig:face-attr-pairs}
\end{figure}

We created the positive and negative pairs carefully for multiple attributes, as shown in Fig.~\ref{fig:face-attr-pairs} . Here, we present the methodology to create such positive and negative pairs. 

\textbf{Face attribute pairs:} We generate the positive and negative pairs for face attribute editing by augmenting the negative source image with the attribute of interest from a source image. Fig. ~\ref{fig:face-attr-pairs} shows examples for all the attribute edits. For attributes such as hat, glasses, bald, bangs, beard, and eye-close, we use the segmentation mask from CelebAMask-HQ~\cite{CelebAMask-HQ} dataset to perform a simple copy-paste operation for the region of interest. For pose, we flipped the source image for positive image creation and for age we use SAM~\cite{yuval2021agetransformation} (state-of-the-art age editing framework) to create an aged face. We use the portrait relighting method ~\cite{portrait-lighting} to get the positive and negative image pairs for lighting. 

Note that one can always use real images of a person at different ages and capture a few images in different lights to obtain such pairs. However, this simple pair creation process fuels the simplicity of our approach and can perform various editing operations that were not possible otherwise. Although the image pairs shown in Fig.~\ref{fig:face-attr-pairs} do not look very realistic, the encoder model maps them to a latent code corresponding to a natural-looking image.  

\begin{figure}[h!]
    \centering
    \includegraphics[width=\linewidth]{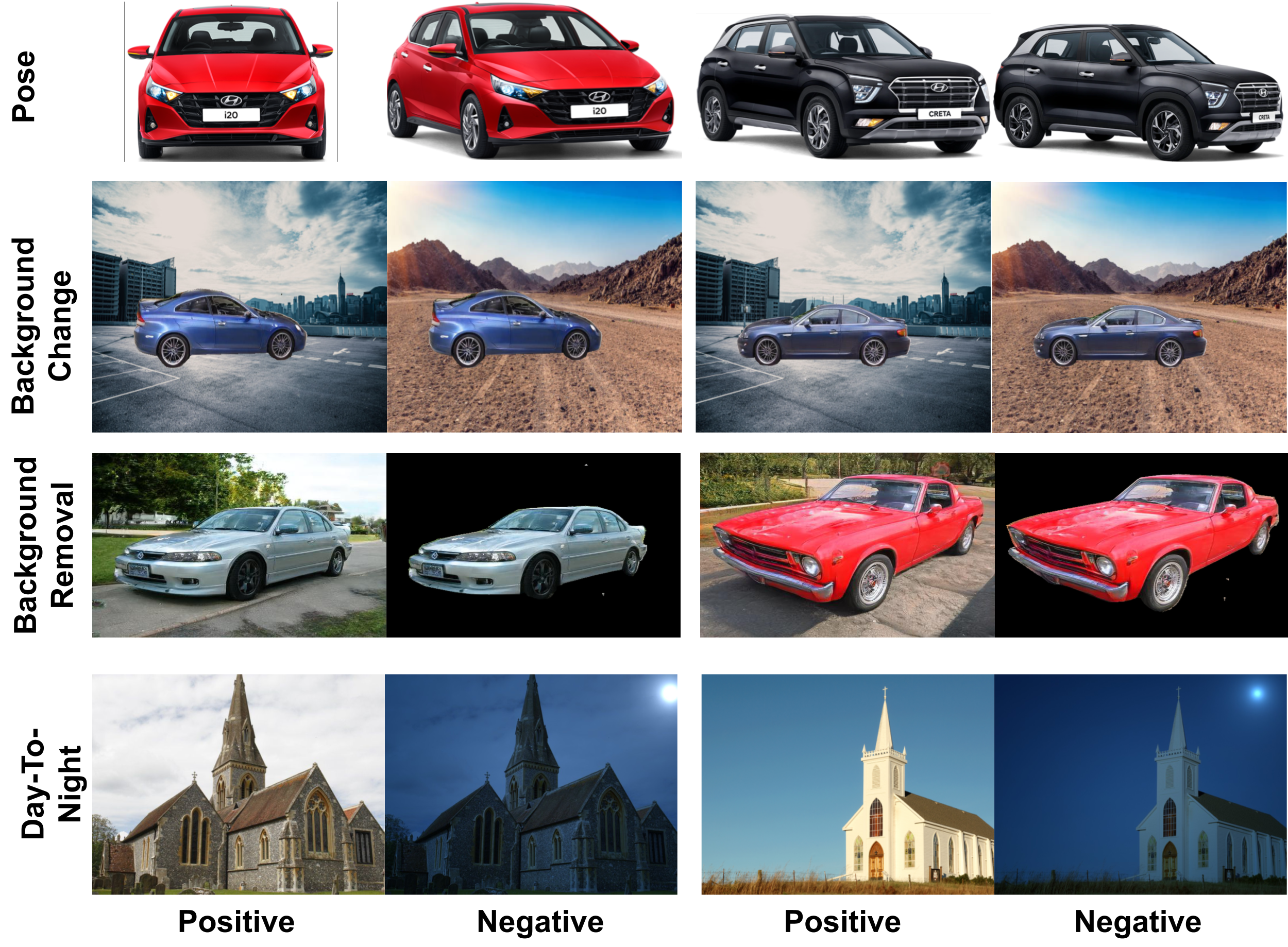}
    \caption{(From Top to Bottom) Sample pairs for Car Pose Change, Car Background Change, Car Background Removal and Church Day-To-Night Task}
    \label{fig:com_car_church_pairs}
\end{figure}

\textbf{Car Pose:} Fig. \ref{fig:com_car_church_pairs} (Row 1) shows positive and negative pairs for the car pose. We generate these pairs through the 360-degree view provided by the car manufacturer on the internet.

\textbf{Car Background Change:} Fig. \ref{fig:com_car_church_pairs} (Row 2) shows positive and negative pairs for the background change task. We use the mask provided or LSUN car images; crop the car from the original image by using the mask and then pasting it on different pre-set backgrounds. Even with this crude way of creating pairs, we achieve excellent results for real car images.

\textbf{Car Background Removal:} Fig. \ref{fig:com_car_church_pairs} (Row 3) shows positive and negative pairs for the background removal task. We use the mask provided for LSUN car images; crop the car from the original image by using the mask and then pasting it on a black background.

\textbf{Church Day-to-Night:} Fig. \ref{fig:com_car_church_pairs} (Row 4) shows positive and negative pairs for the church day-to-night task. We use an online tool that converts day images to night by applying a gamma correction to the original image. 



\section{Additional Results on other datasets}
\label{sec:add_res}
\begin{figure}[h!]
\includegraphics[width=0.40\textwidth]{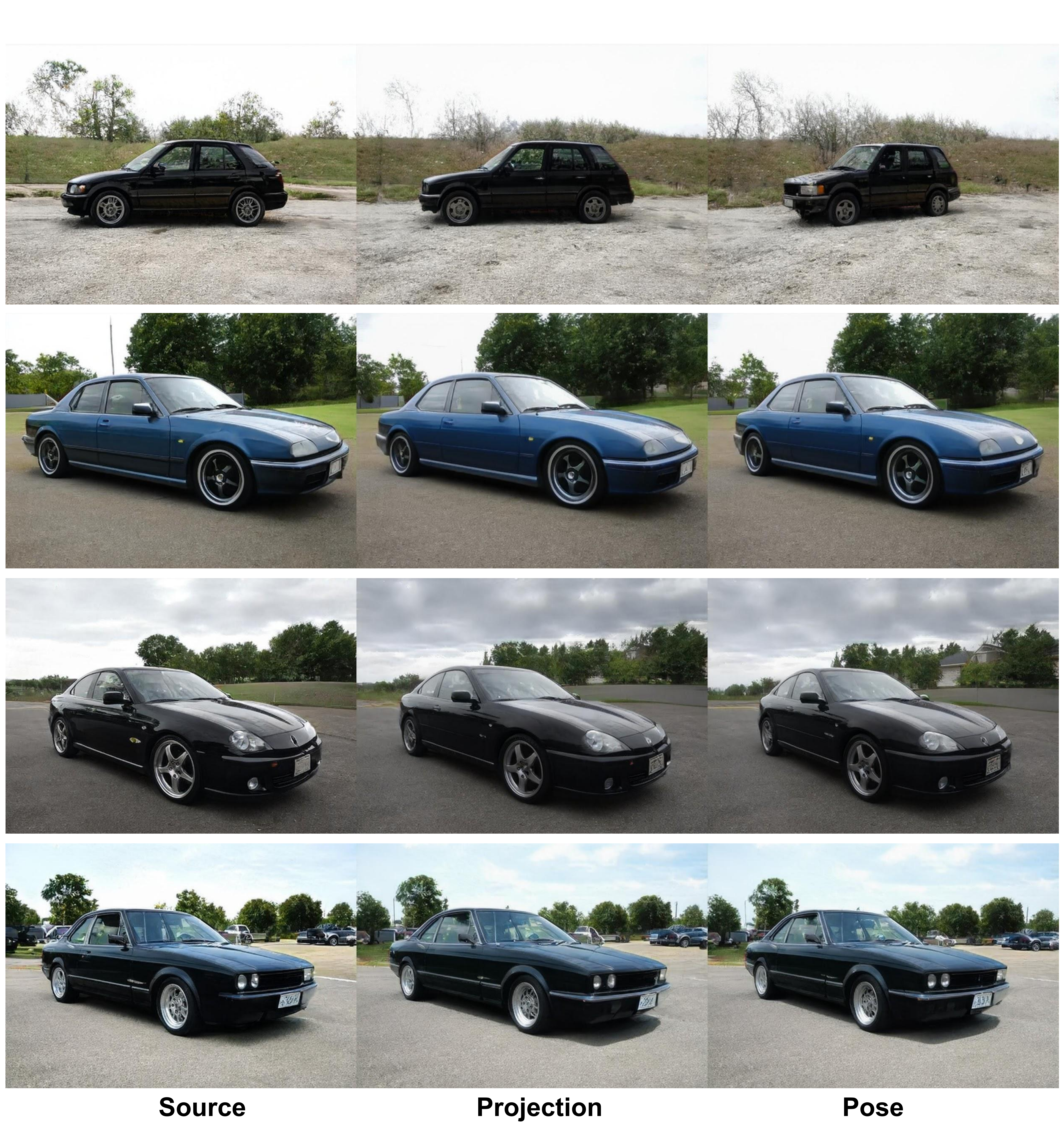} 
\caption{Editing for car pose using FLAME}  
\label{fig:car_pose}
\end{figure}

\begin{figure}[h!]
\includegraphics[width=0.40\textwidth]{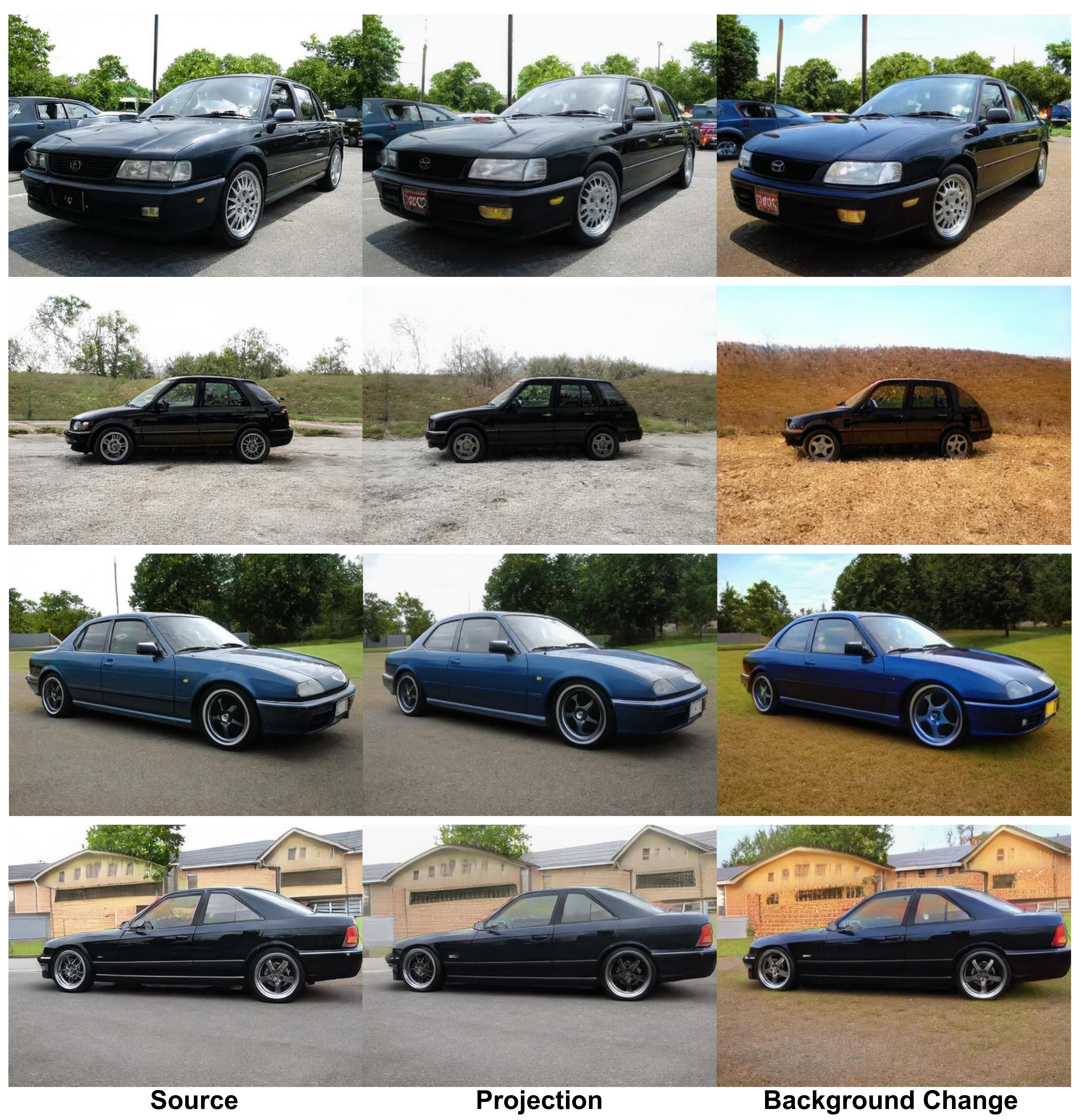}
\caption{Editing car background using FLAME}  
\label{fig:car_bgd_change}
\end{figure}

\begin{figure}[h!]
\includegraphics[width=0.40\textwidth]{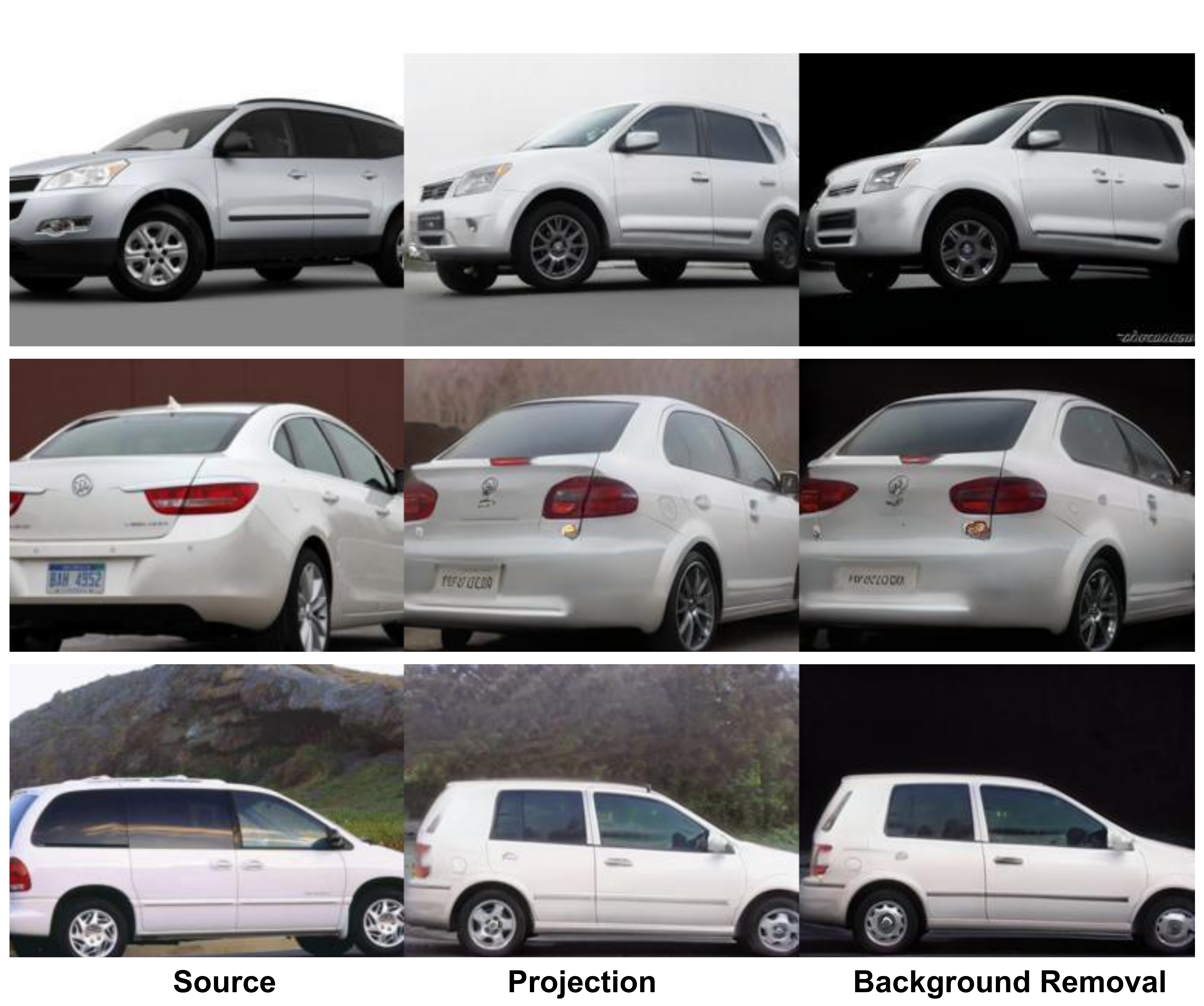}
\caption{Removal of background using FLAME}  
\label{fig:car_bgd_rem}
\end{figure}

\begin{figure}[!t]
\includegraphics[width=0.40\textwidth]{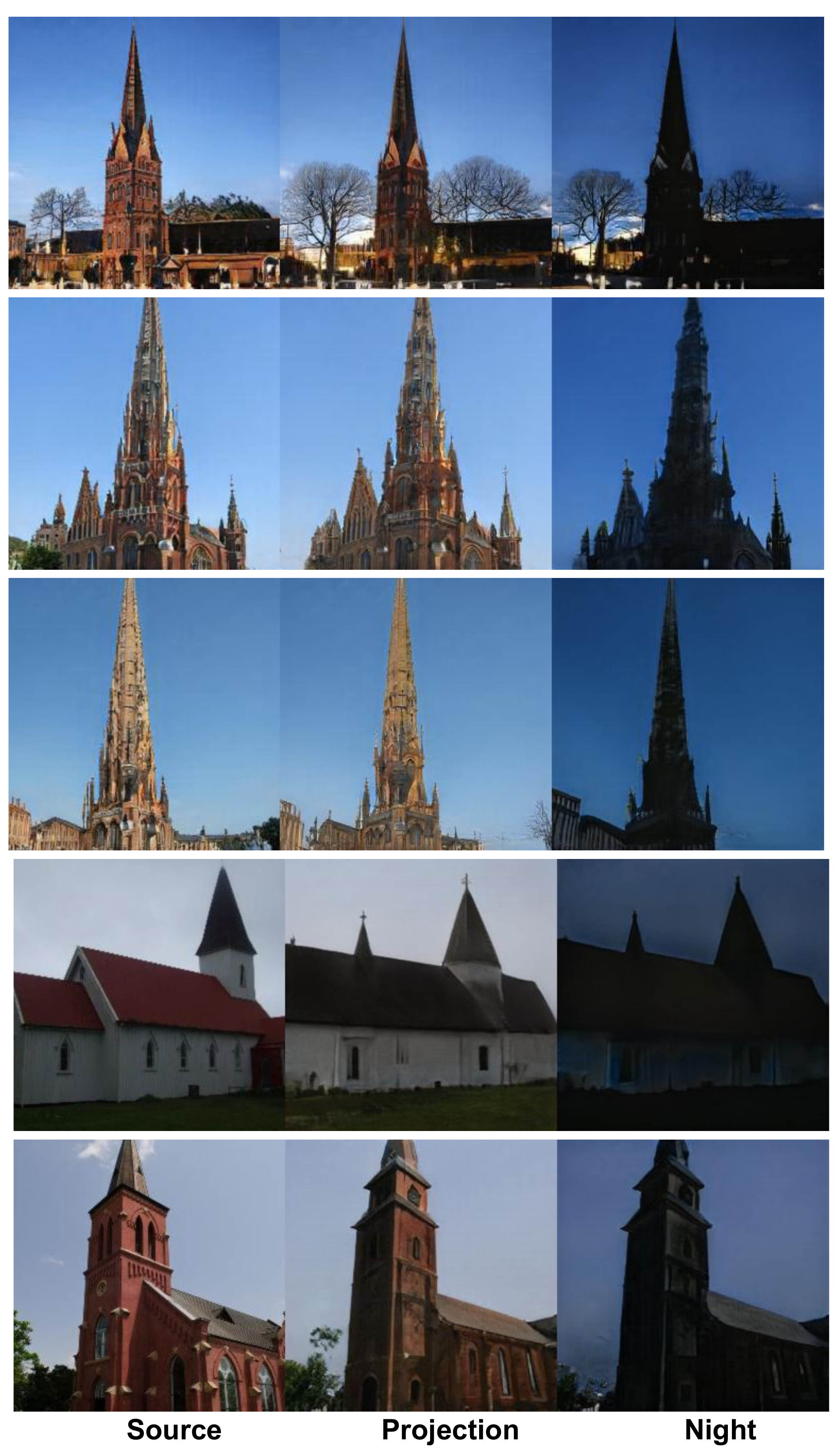}
\caption{Changing the scene from day to night using FLAME}  
\label{fig:church_night}
\end{figure}

To evaluate the generalization ability of FLAME for editing on other categories, we performed background change (Fig. ~\ref{fig:car_bgd_change}), background removal (Fig. ~\ref{fig:car_bgd_rem}) and pose change (Fig. ~\ref{fig:car_pose}) for car dataset and day to night change (Fig. ~\ref{fig:church_night}) for churches dataset. We can observe in all these edits that FLAME is able to disentangle the attribute edits and results in realistic-looking edits. 

\begin{figure*}[h!] 
\includegraphics[width=0.90\textwidth]{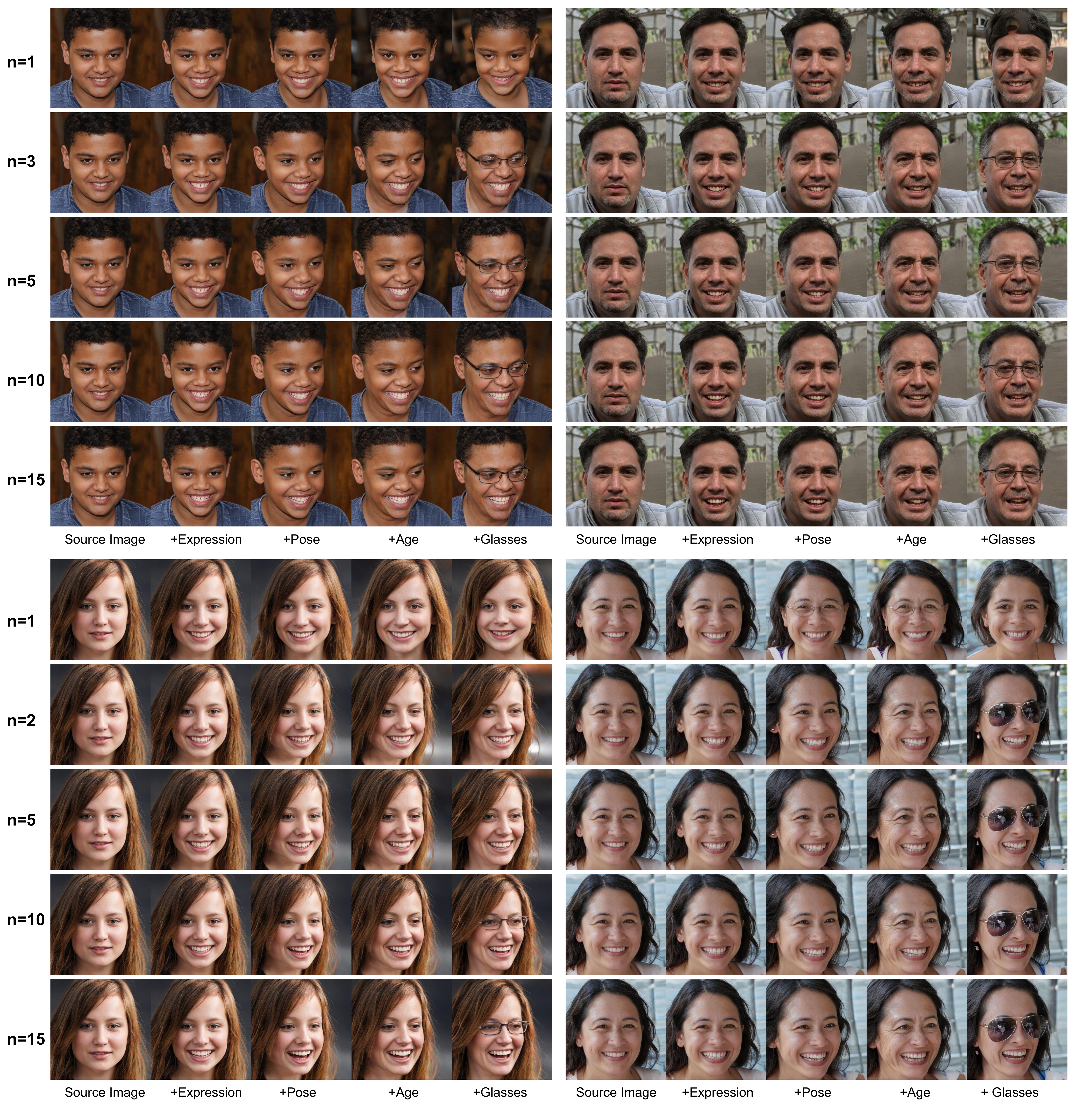}
\caption{Ablation study with number of synthetic image pairs used for semantic direction estimation.} 
\label{fig:ablation}
\end{figure*}



\section{Ablation study on number of image pairs used for direction estimation}
\label{sec:num_imgs}

Here we perform an ablation study on the number of synthetic positive-negative image pairs $n$ used to estimate the latent space's edit direction. Fig.~\ref{fig:ablation} show the results. We observe that taking only one image pair is not sufficient as it often results in editing multiple attributes at once and is also not accurate, e.g., changing the pose attribute in another direction. Also, we observe that by increasing the number of image pairs, we can find realistic edit operations for the attributes while preserving other attributes. One interesting observation is that irrespective of the number of pairs used, our approach preserves the identity of the faces. This can be attributed to the utility of synthetic image pairs in estimating the directions.


\begin{figure}[!b]
\includegraphics[width=0.5\textwidth]{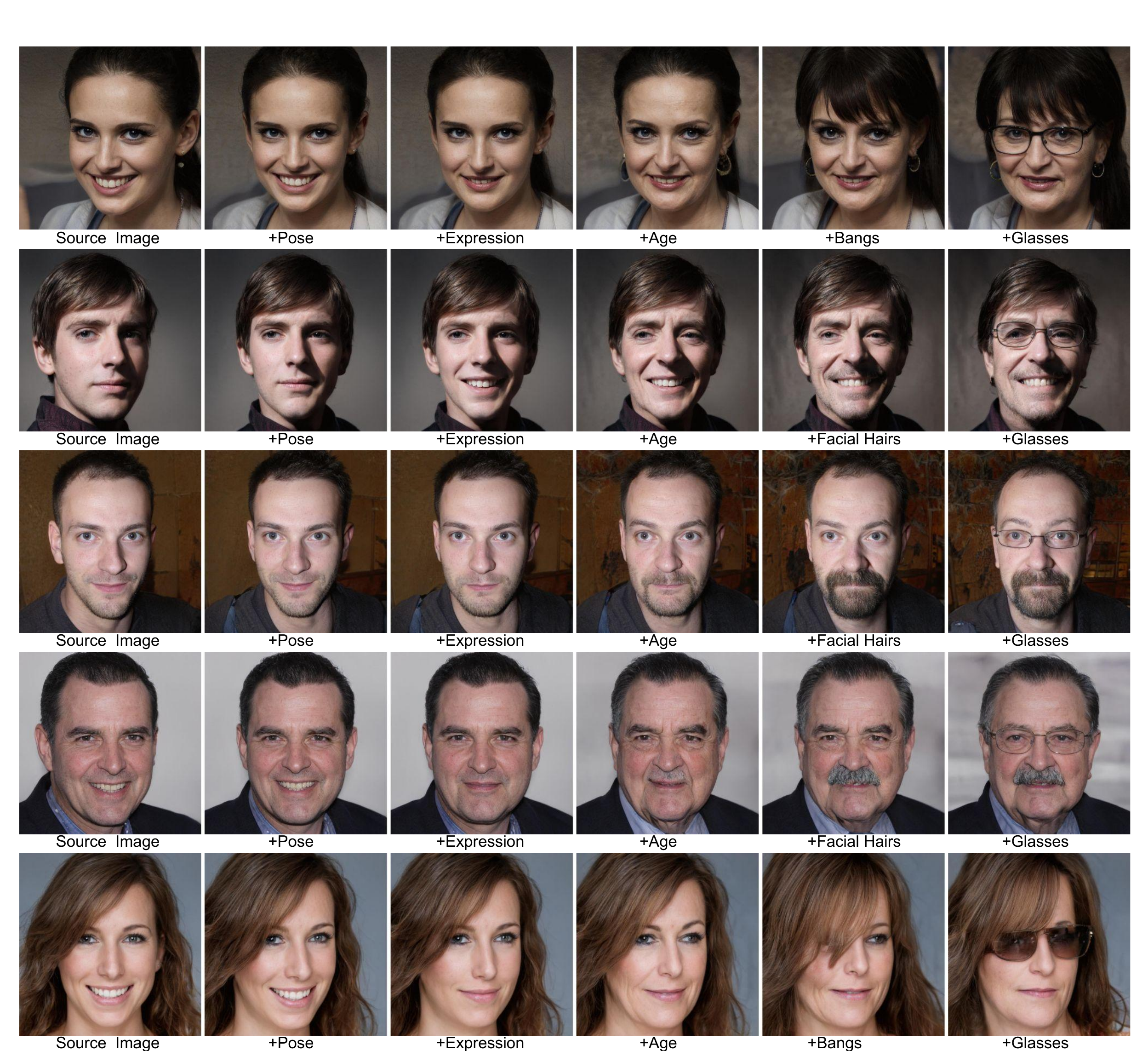}
\caption{Sequential edit on synthetic face images} 
\label{fig:synthetic1}
\end{figure}

\begin{figure}[!t]
\includegraphics[width=0.5\textwidth]{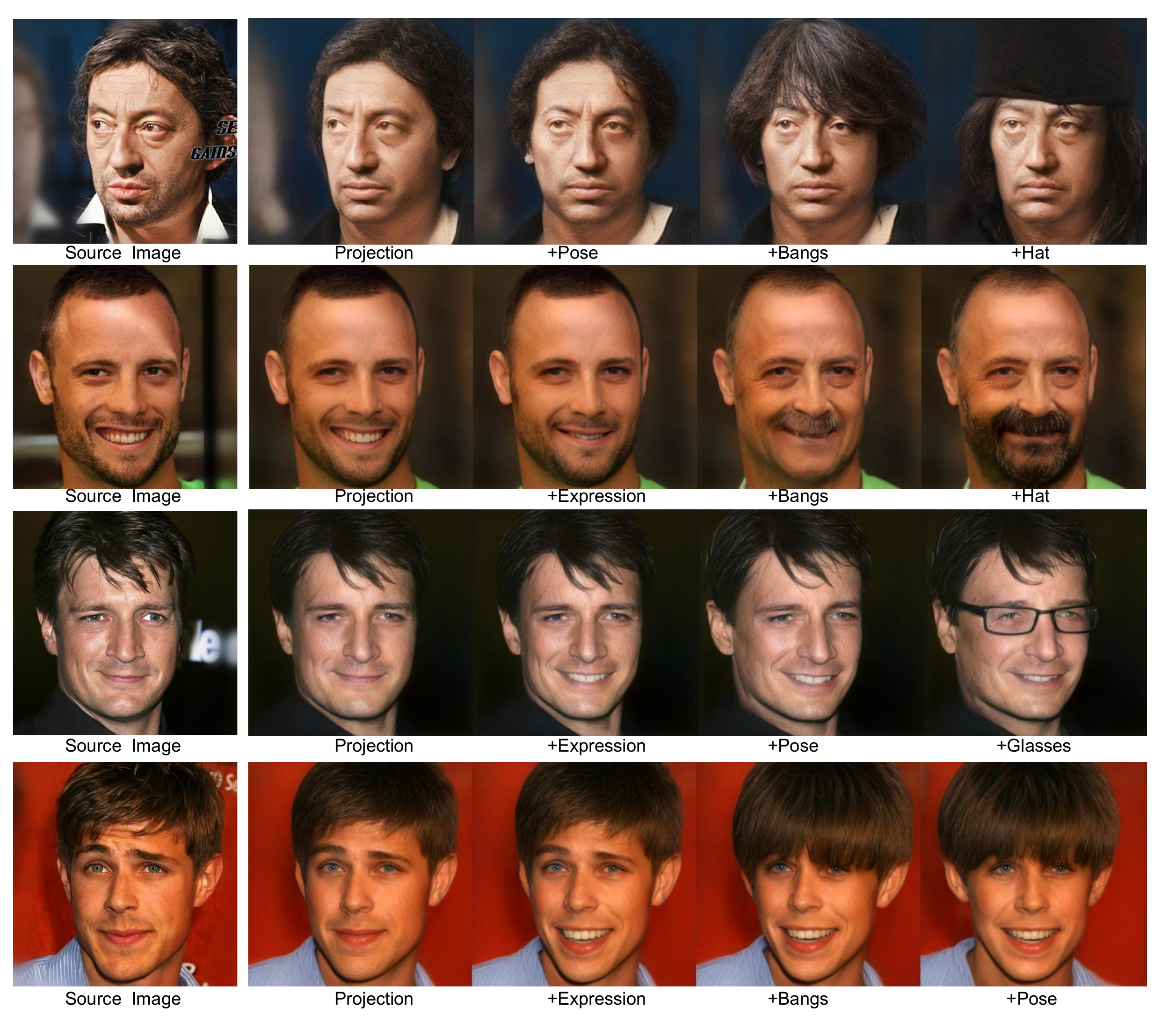}
\caption{Sequential edit on real images}
\label{fig:real1}
\vspace{5mm} 
\end{figure}

\section{Additional results for sequential attribute editing}
\label{sec:seq_ed}
Fig.~\ref{fig:synthetic1} show the results for editing on synthetic dataset. We can observe that our method generates diverse image edit operations with very high fidelity. Also, note that our method preserves the identity for the source image even after performing multiple sequential edits.  
Additionally, we present result for attribute editing on real images by first projecting them into the latent space using a StyleGAN2 encoder model and performing latent transformation for attribute editing. Fig. ~\ref{fig:real1} shows results from this section. We observe that our method generates realistic edits while preserving a given subject's identity and other attributes.

\begin{figure*}[!t]
\includegraphics[width=0.8\textwidth]{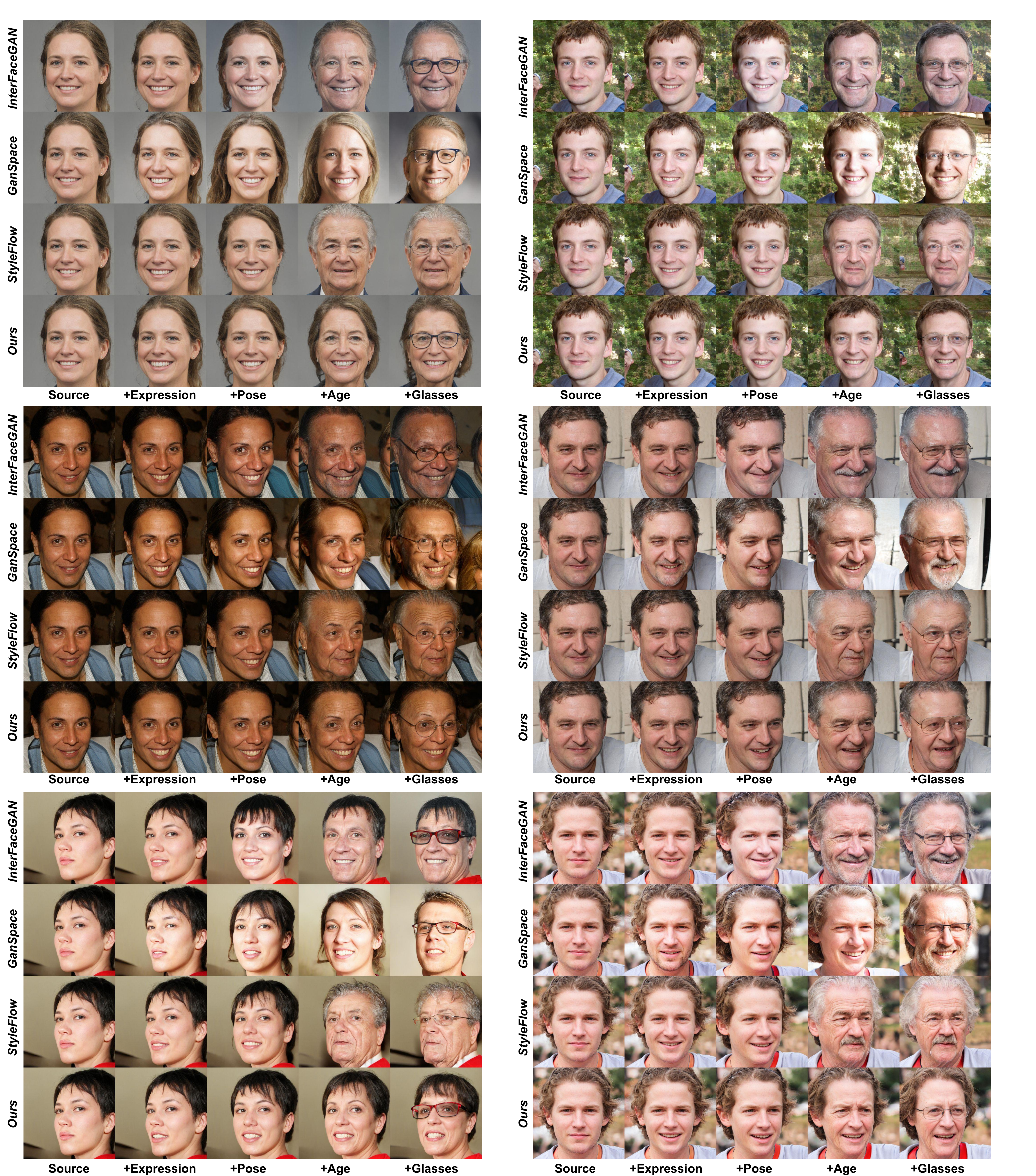}
\caption{Comparison with State-of-the-Art Methods InterFaceGAN, GANSpace, StyleFlow on sequential image editing operations}
\label{fig:comparison}
\end{figure*}


\section{Comparison Results for Attribute Editing}
\label{sec:comparison}
This section shows additional comparison results with state-of-the-art methods GANSpace, InterFaceGAN, and StyleFlow. We present results for sequential image edits on synthetic images in Fig. ~\ref{fig:comparison}. Note that our proposed method preserves identity in all the cases and performs disentangled attribute editing operations. Also, our method preserves the color tone and lighting of the scene in all of our sequential editing results. 

\begin{figure*}[h!]
\includegraphics[width=\textwidth]{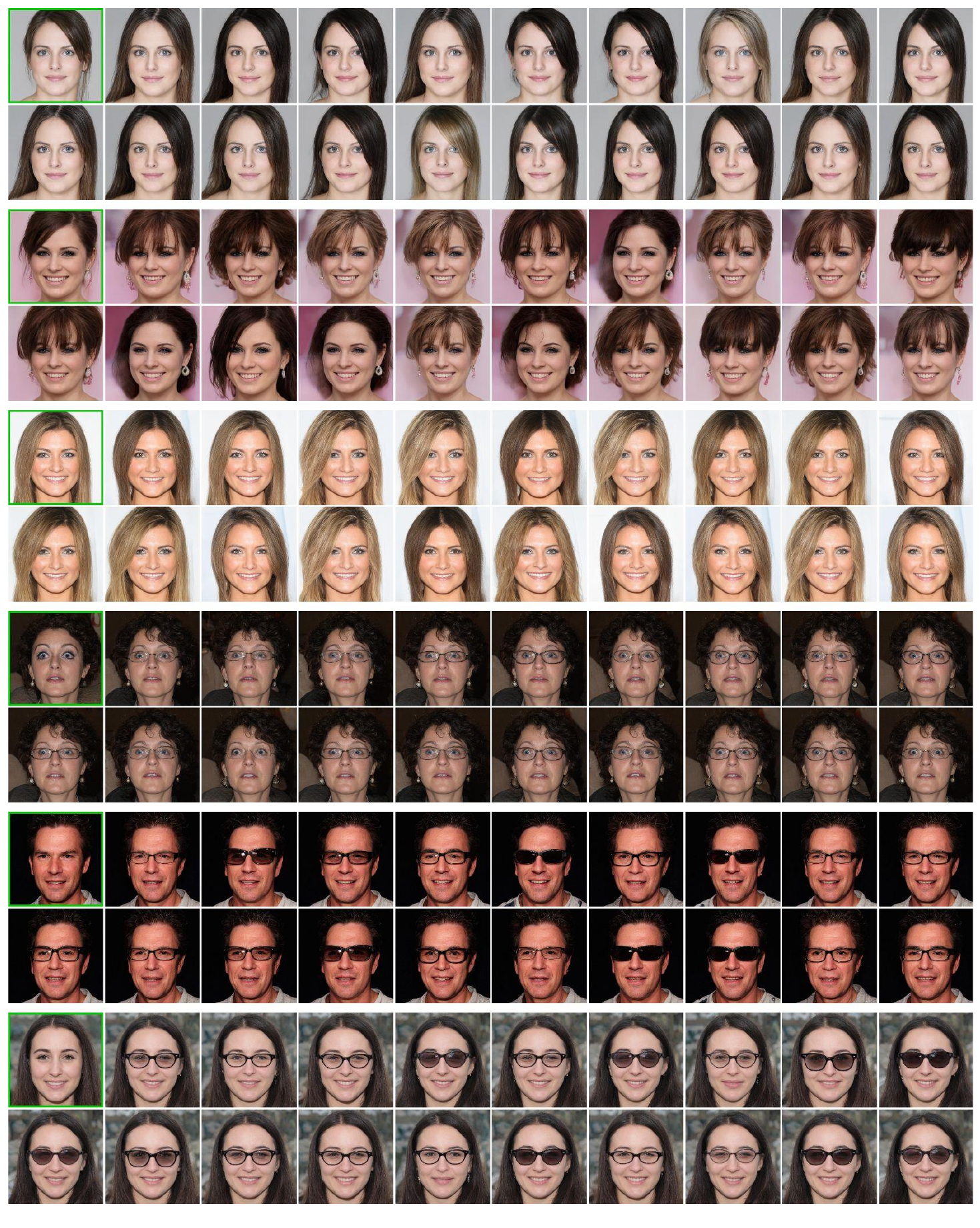}
\caption{Results for attribute styles for hairs and glasses}
\label{fig:att-style}
\end{figure*} 

\section{Additional Results for attribute style manipulation} 
In this section, we present additional results for attribute style manipulation. We have shown results for hair-style and eye-glass variations for $3$ input examples each in Fig. ~\ref{fig:att-style}. Our approach generates diverse attribute styles both for eyeglasses and hairstyles. Our approach preserves other facial attributes during attribute style manipulation, including lighting and head pose. These results efficacies our hypothesis that the obtained latent-space manifold is largely disentangled and controls styles of a single attribute without altering other attributes. Such a framework opens up a great opportunity for synthetic data creation for downstream tasks such as face recognition.

\section{Semantic Direction Estimation}
\label{sec:dir_est}
Here we explain how to solve the following optimization problem to obtain a closed-form solution.

\begin{equation}
\label{eq:1}
    \mathbf{\hat{d_j}} = argmax_{\mathbf{d_j}, |\mathbf{d_j}| = 1} \sum_{k=1}^{n}\left<\mathbf{d^k_j}, \mathbf{d_j}\right>^2
\end{equation}

We want to estimate an r-dimensional vector $d_j$ such that its sum of squared dot product with all the $n$ vectors $\mathbf{d^k_j}$ is maximized. We stack all the vectors $\mathbf{d^k_j}$  as rows to form a matrix $\mathbf{A}$ of dimension $n*r$. The Singular Value Decomposition for matrix $\mathbf{A}$ is given by: 

\begin{equation}
\label{eq:2}
    \mathbf{A} = \mathbf{U}\mathbf{\Sigma}\mathbf{V^T}
\end{equation}    
where columns of $V$ are the orthonormal basis for $r$ dimensional space and sigma is a diagonal matrix.

The optimization problem in Eq ~\ref{eq:1} can be rewritten as:

\begin{equation}
\label{eq:3}
    \mathbf{\hat{d_j}} = argmax_{\mathbf{d_j}, |\mathbf{d_j}| = 1} ||\mathbf{A} \mathbf{d_j}||^2 
\end{equation}


We can re-write any vector $\mathbf{d_j}$ in the following form as $v_j's$ form the basis of $r$ dimensional space:

\begin{equation}
\label{eq:4}
    \mathbf{d_j} = \sum_i \left<\mathbf{d_j}, \mathbf{v_i}\right>\mathbf{v_i}
\end{equation}

\begin{equation}
\label{eq:5}
    ||\mathbf{A}\mathbf{d_j}||^2 = ||\mathbf{A} \sum_i \left<\mathbf{d_j}, \mathbf{v_i}\right>\mathbf{v_i}||^2
\end{equation}

\begin{equation}
\label{eq:6}
    ||\mathbf{A}\mathbf{d_j}||^2 = ||\sum_i \left<\mathbf{d_j}, \mathbf{v_i}\right>\mathbf{A}\mathbf{v_i}||^2
\end{equation}

Using $\mathbf{A}\mathbf{v_i}$ = $\sigma_i \mathbf{u_i}$, since $u$ and $v$ are coming from SVD of A, Eq. ~\ref{eq:5}. 

\begin{equation}
\label{eq:7}
    ||\mathbf{A}\mathbf{d_j}||^2 = ||\sum_i \left<\mathbf{d_j},\mathbf{v_i}\right>\sigma_i \mathbf{u_i}||^2
\end{equation}

As all the vectors $\mathbf{d_j}$, $\mathbf{u_i}$ and $\mathbf{v_i}$ are of unit norm, the above equation can be maximized when $\mathbf{d_j}$ is equal to the vector $\mathbf{v_{max}}$ corresponding to the maximum singular value $\mathbf{\sigma_{max}}$. Hence, the eigen vector $\mathbf{v_{max}}$ corresponding to the maximum eigen value will be a closed form solution of Eq. ~\ref{eq:1}. 

\clearpage
 


\end{document}